\documentclass[journal]{IEEEtran}
\usepackage{amsfonts}
\usepackage{amssymb}
\usepackage{hyperref}
\usepackage{graphicx}
\usepackage{enumerate}
\usepackage{amsmath}
\usepackage{color}
\usepackage{amsthm}
\usepackage{verbatim}
\usepackage{amsmath}
\usepackage{algorithm}
\usepackage{cite}
\usepackage{algpseudocode}
\usepackage{subfig}
\newtheorem{assumption}{Assumption}

\newtheorem{theorem}{Theorem}

\usepackage{graphicx}
\usepackage{float}
\usepackage{autobreak}
\usepackage{amsfonts,amssymb}
\usepackage{enumitem}
\usepackage[fontsize=10pt]{fontsize}
\allowdisplaybreaks

\ifCLASSINFOpdf
\else
\fi
\begin{document}
%
\title{Denoising and Adaptive Online Vertical Federated Learning for Sequential Multi-Sensor Data in Industrial Internet of Things}

\author{{Heqiang Wang, Xiaoxiong Zhong, Kang Liu, Fangming Liu, Weizhe Zhang} 
\thanks{
H. Wang, X. Zhong, K. Liu, F. Liu and W. Zhang are with Department of New Networks, Peng Cheng Laboratory, Shenzhen, 518066, China.} 
\thanks{Corresponding Authors: Xiaoxiong Zhong, Kang Liu.}
}


%


\maketitle

\begin{abstract}
With the continuous improvement in the computational capabilities of edge devices such as intelligent sensors in the Industrial Internet of Things, these sensors are no longer limited to mere data collection but are increasingly capable of performing complex computational tasks. This advancement provides both the motivation and the foundation for adopting distributed learning approaches. This study focuses on an industrial assembly line scenario where multiple sensors, distributed across various locations, sequentially collect real-time data characterized by distinct feature spaces. To leverage the computational potential of these sensors while addressing the challenges of communication overhead and privacy concerns inherent in centralized learning, we propose the \underline{D}enoising and \underline{A}daptive \underline{O}nline Vertical Federated Learning (DAO-VFL) algorithm. Tailored to the industrial assembly line scenario, DAO-VFL effectively manages continuous data streams and adapts to shifting learning objectives. Furthermore, it can address critical challenges prevalent in industrial environment, such as communication noise and heterogeneity of sensor capabilities. To support the proposed algorithm, we provide a comprehensive theoretical analysis, highlighting the effects of noise reduction and adaptive local iteration decisions on the regret bound. Experimental results on two real-world datasets further demonstrate the superior performance of DAO-VFL compared to benchmarks algorithms.
\end{abstract}

\begin{IEEEkeywords}
Industrial Internet of Things, Vertical Federated Learning, Online Learning, Deep Reinforcement Learning 
\end{IEEEkeywords}

%
\IEEEpeerreviewmaketitle

\section{Introduction}
Recent advancements in communication technologies and the proliferation of intelligent edge devices, coupled with the rapid pace of industrial informatization, have driven the widespread adoption of the Industrial Internet of Things (IIoT) \cite{boyes2018industrial}. This growth is largely attributed to IIoT's potential to significantly enhance productivity and efficiency across various industries. As we move forward into future industrial revolutions, particularly Industry 4.0 \cite{lu2017industry}, the IIoT is expected to play a pivotal role in the development of new applications such as smart manufacturing, smart factories, smart transportation, and smart healthcare. To enable intelligent services and applications within the IIoT ecosystem, artificial intelligence (AI) techniques, particularly machine learning (ML) and deep learning (DL), are widely used to train models on industrial data. Traditionally, this training has been conducted in centralized cloud environments or data centers. However, this approach faces significant challenges as IIoT data volumes continue to grow. The need to transfer large volumes of IIoT data to centralized servers for model training demands substantial network bandwidth and introduces considerable communication overhead, which is impractical for time-sensitive IIoT applications like autonomous driving \cite{levinson2011towards} and real-time healthcare \cite{al2018context}. Moreover, uploading sensitive data to the cloud increases the risk of privacy breaches. In response to these challenges, distributed learning based on edge devices, particularly federated learning (FL) \cite{lim2020federated}, has gained attention as a promising solution. FL offers a more cost-effective and privacy-preserving alternative for deploying intelligent IIoT applications in a distributed manner, minimizing the need for data transfer while ensuring that sensitive information remains processed locally.

\begin{figure}[htp]
\vspace{-5pt}
\centering
\subfloat{\includegraphics[width=0.9\linewidth]{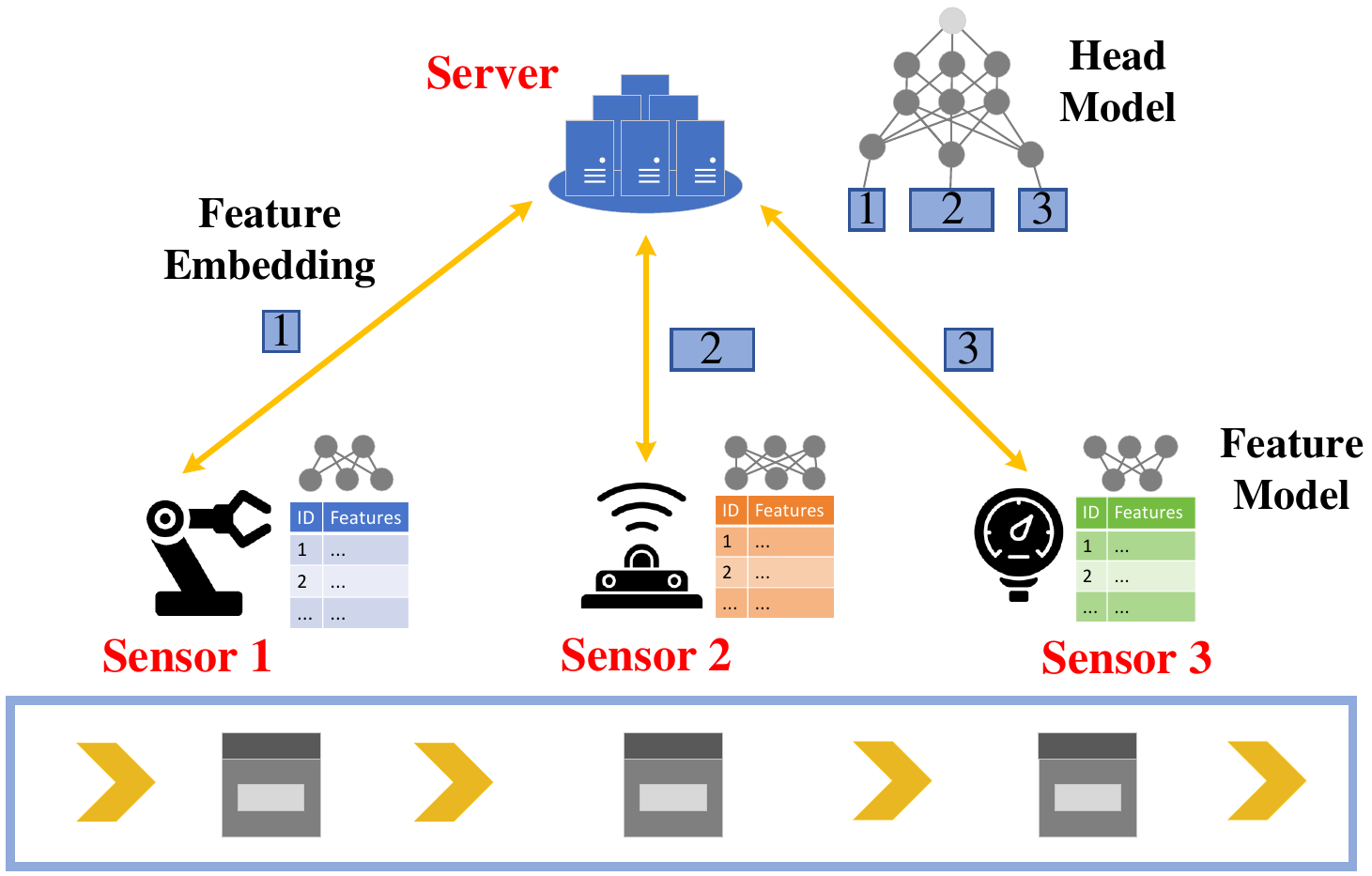}} 
\caption{DAO-VFL for IIoT-Based Assembly Line}  
\label{Dao_vfl}
\vspace{-10pt}
\end{figure}

The IIoT encompasses a broad array of application scenarios, each presenting its own set of problems, challenges, and potential solutions.  In this work, we focus on an industrial assembly line scenario, as illustrated in Fig.~\ref{Dao_vfl}, where multiple sensors collaborate for tasks like quality control and fault detection. As sensors become increasingly advanced, their roles have expanded beyond simple data collection and transmission to encompass more complex tasks such as model training and inference. In industrial assembly lines, numerous sensors are distributed along the line. Due to variations in the types and locations of these sensors, the data they collect for a product moving through the assembly line can be seen as different features of the same dataset. This setup aligns perfectly with the concept of vertical federated learning (VFL) \cite{liu2024vertical, wei2022vertical}, where participants have distinct feature spaces but share the same set of data samples. The core idea behind VFL is to partition the DNN into distinct segments, each trained separately by individual sensors and the server, ensuring that raw data remains at the sensors. The DNN is divided into a head classifier and multiple vertically separable feature extractors, one for each sensor. Instead of transmitting raw data to the server, each sensor processes its local data through its own feature extractor, generating a feature embedding, which is then sent to the server. The server's head classifier then processes the feature embeddings from all the sensors to derive the final results. 

In addition to adopting the VFL framework described above, there are three key challenges that must also be addressed in an IIoT-based assembly line scenario. First, since sensor data in industrial assembly line is collected in real-time and may not be fully available before training starts, the algorithm must be designed using an online learning approach, unlike traditional offline learning methods that rely on a static dataset. Second, communication between sensors over wireless networks in industrial environments often experiences interference and is subject to various levels of noise \cite{liu2020noise}. Therefore, it is critical to develop effective noise reduction techniques to mitigate these effects, which would enhance the overall learning performance. Third, considering the varying computational capabilities and network conditions of sensors in industrial assembly lines \cite{wu2019online}, and in order to achieve synchronized learning, it is crucial to design algorithms that determine the appropriate number of local iterations for each sensor. This approach helps reduce overall latency while enhancing learning performance.

Building on the challenges discussed above, this work introduces the concept of Denoising and Adaptive Online Vertical Federated Learning (DAO-VFL) within an IIoT based assembly line scenario. Unlike previous VFL studies, our approach directly addresses practical issues encountered in IIoT environments. It emphasizes the integration of \underline{D}enoising \underline{A}daptive and \underline{O}nline mechanisms within the VFL framework, which have not been systematically explored in prior VFL research. Our primary contributions can be summarized as follows:
\begin{enumerate}
    \item We formulate the multi-sensor distributed learning problem in IIoT-based assembly lines as an online VFL problem. Building on this problem, and addressing challenges such as communication noise and sensor heterogeneity in industrial environments, we propose the DAO-VFL algorithm, which can facilitate practical online training by incorporating noise reduction techniques and adaptive local iteration decisions. 
    \item We conduct a comprehensive theoretical analysis of the DAO-VFL algorithm, accounting for the impacts of communication noise and adaptive local iteration decisions. The derived regret bound highlights these impacts and provides insights into selecting appropriate local iteration decisions for each sensor.
    \item To achieve noise reduction, we incorporated a denoising autoencoder seamlessly into the training process. To determine the local iteration decisions for each sensor, we formulated an optimization problem that accounts for learning performance, overall latency, and local iteration disparities. This problem is solved using a deep reinforcement learning approach, to derive the local iteration decisions for each sensor at every global round.
    \item In the experimental section, we evaluated the DAO-VFL algorithm using two datasets: the widely used CIFAR-10 dataset and the real-world IIoT-based C-MAPSS dataset for residual life prediction. Our results not only assess the overall performance of the DAO-VFL algorithm but also provide a detailed analysis of its noise reduction capabilities and adaptive local iteration decision-making mechanisms.
\end{enumerate}

The rest of this paper is organized as follows. Section II reviews related works on FL for IIoT, VFL, and online FL. Section III introduces the system model and problem formulation. In Section IV, we introduce the specific process of DAO-VFL algorithm. Section V presents the regret analysis of DAO-VFL. Section VI formulates and addresses the optimization problem for determining the local iterations decision for each sensor. The experimental results of DAO-VFL are presented in Section VII. Finally, Section VIII concludes the paper.

\section{Related Work}
\subsection{FL for Industrial Internet of Things}
As the IIoT continues to evolve, the demand for increased intelligence within IIoT is becoming increasingly urgent. AI techniques, particularly DL, play a crucial role in enabling this intelligence and are widely utilized in IIoT environments. However, the inherent characteristics of IIoT, such as the large number of distributed computing nodes and complex network architectures \cite{sisinni2018industrial}, pose significant challenges to traditional centralized learning approaches, making them inadequate for these scenarios. To address these challenges, FL has emerged as a promising alternative, gaining significant attention for its ability to better align with the unique characteristics of IIoT systems. The works of \cite{nguyen2021federated, boobalan2022fusion} provide comprehensive overviews of FL for IIoT. Unlike conventional FL, IIoT-based FL is typically tailored to specific application domains, such as smart manufacturing \cite{kevin2021federated}, smart transportation \cite{xu2022efficient} and smart grid \cite{su2021secure}. Additionally, IIoT-based FL often addresses critical and urgent industrial issues, including anomaly detection \cite{wang2021toward} and intrusion detection \cite{mao2024towards, zhang2022federated}. However, research in IIoT-based FL remains in its early stages, with most current studies focusing on ideal conditions and overlooking critical challenges specific to IIoT environments. These challenges include noise in the industrial environment \cite{liu2020noise}, limited data storage capacity of edge devices \cite{wang2023local}, and the heterogeneity of IIoT edge devices \cite{wu2019online}. In this work, we systematically investigate FL deployment within IIoT-based assembly line scenarios. We aim to address and provide solutions to some of the critical challenges posed by real industrial environments, thereby advancing the practical applicability of FL in IIoT environment.

\subsection{Vertical Federated Learning}
In recent years, VFL has garnered significant attention. Originally introduced by \cite{hardy2017private}, VFL operates on vertically partitioned data, setting it apart from the concept of horizontal FL (HFL)\cite{zhu2021federated}. Comprehensive surveys such as \cite{liu2024vertical, wei2022vertical} have further expanded on the scope of VFL. Unlike horizontal FL, VFL faces its own set of unique challenges. Some studies \cite{feng2022vertical, kang2022fedcvt, fan2022fair} have focused on optimizing data utilization to enhance the effectiveness of the joint model in VFL. Other research efforts \cite{sun2022label, sun2021defending} have concentrated on creating privacy-preserving protocols to mitigate the risks of data leakage. Additionally, efforts have been made to reduce communication overhead, either by incorporating multiple local updates per iteration \cite{castiglia2023flexible, zhang2022adaptive} or by employing data compression techniques \cite{castiglia2022compressed, wang2024computation}. VFL's practical benefits, particularly in enabling data collaboration among diverse institutions across various industries, have heightened interest from both academic and industrial communities. VFL has found applications in a wide range of fields, including recommendation systems \cite{cui2021exploiting}, finance \cite{ou2020homomorphic}, and healthcare \cite{chen2020vafl}. The VFL approaches outlined in the previous section exhibit two significant limitations. First, they primarily focus on offline learning with static datasets, overlooking the dynamic nature of stream data. Second, they do not adequately account for or address the unique challenges inherent to real IIoT scenarios. These limitations form the core issues that this work seeks to resolve.

\subsection{Online Federated Learning}
Online learning is tailored to process data sequentially and update models incrementally, making it particularly effective for applications where data continuously arrives and models need to adapt to new patterns in real-time \cite{sahoo2017online}. These methods are computationally efficient and have the distinct advantage of not needing the entire dataset to be available at the outset, making them ideal for scenarios with limited memory resources. Consider the FL scenario, online federated learning (OFL) \cite{hong2021communication} has emerged as an innovative paradigm that extends the principles of online learning to a network of multiple learners or agents. The key difference between OFL algorithms and traditional FL algorithms lies in the goal of local updates. While traditional FL algorithms aim to find a single global model that minimizes a global loss function, OFL algorithms focus on identifying a sequence of global models that minimize cumulative regret. Recent years have seen limited research on OFL. For instance, the authors of \cite{kwon2023tighter} propose a communication-efficient OFL method that balances reduced communication overhead with strong performance. Similarly, \cite{mitra2021online} introduces FedOMD, an OFL algorithm tailored for uncertain environments, capable of processing streaming data from devices without relying on statistical assumptions about loss functions. While the aforementioned studies primarily focus on the HFL scenario, work \cite{wang2023online} addresses the VFL context by proposing an online VFL architecture tailored to cooperative spectrum sensing problems, achieving a sublinear regret rate. However, these studies are based on idealized scenarios and do not address the challenges of applying online VFL in real industrial scenarios, such as noise interference and device heterogeneity.

\section{System Model}
Consider an IIoT assembly line system consisting of a single server and $K$ smart sensors. Each sensor collects distinct system parameters along the assembly line, such as temperature, pressure, humidity, as illustrated in Fig.~\ref{Dao_vfl}. These parameters are treated as independent features within a single data sample, reflecting the unique identities and locations of each sensor. As a result, the distributed training process involving these sensors falls under the category of a VFL problem, since each sensor operates within a distinct feature space. During the operation of the assembly line, the sensors continuously gather new data over time, with the timeline divided into discrete periods represented as $t = 1, 2, ..., T$.

In each time period,  each sensor $k \in \mathcal{K}$ collects a local training dataset consisting of $N$ data samples, represented as $\textbf{x}^t_k \in \mathbb{R}^{N \times P_k}$, where $P_k$ is the dimension of the raw data collected by sensor $k$. The individual data samples, denoted as $x^{t, n}_k$ for all sensors, are gathered simultaneously and linked to a common label, $y^{t, n}$, which indicates, for instance, the product's conformity. The collective training dataset at period $t$ is denoted as $\textbf{x}^t \in \mathbb{R}^{N \times P}$, where $P = \sum_{k=1}^K P_k$. It should be emphasized that although $\textbf{x}^t$ represents the entirety of data collected at a given period, these data are gathered and utilized locally by each sensor individually and are not uploaded to the server during the training process.

In the VFL framework, each sensor trains a distinct local feature model parameterized by $\theta_k$ to process its collected raw data. Concurrently, the server trains a server head model represented by the parameter $\theta_0$. The combined parameters of the entire model are denoted as $\Theta = [\theta^\top_0, \theta^\top_1, ..., \theta^\top_K]^\top$. The raw data $x^{t,n}_k$ collected by each sensor is processed through its local feature model, a process referred to as feature extraction, and represented as $h_k(\theta_k; x^{t,n}_k)$. This operation transforms high-dimensional raw data into low-dimensional feature representations, which facilitate the learning process of the server head model. Based on these definitions, the loss function for the collective training dataset at period $t$ is expressed as follows: 
\begin{align}
F_t(\Theta; \textbf{x}^t, \textbf{y}^t)  = \frac{1}{N}\sum_{n=1}^N l_t(\theta_0, \{h_k(\theta_k; x^{t,n}_k)\}_{k=1}^K; y^{t,n})
\end{align}
where $l_t(\cdot)$ represents the loss function for the single data sample. For simplicity, the feature embedding of the dataset is denoted as $\textbf{x}^{t}_k$ as $h_k(\theta_k; \textbf{x}^t_k)$, which is often abbreviated as $h_k(\theta_k; \textbf{x}^t_k) = h^t_k(\theta_k)$. Additionally, we assign $k = 0$ to the server, defining $h_0(\theta_0) = \theta_0$. In this context, $h_0(\theta_0)$ refers to the head model rather than the feature embedding. Furthermore, the overall loss function is expressed as $F_t(\Theta) = F_t(h_0(\theta_0), h_1(\theta_1), ..., h_K(\theta_K))$.

Since the training process relies on a dynamically collected, real-time data instead of a static dataset, an online learning approach becomes essential. Let the overall model at each period be represented as $\Theta^1, \ldots, \Theta^T$. The learning regret, $\text{Reg}_T$ is defined to quantify the gap between the cumulative loss incurred by the learner and the cumulative loss of an optimal fixed model in hindsight. Specifically:
\begin{align}
  \text{Reg}_T = \sum_{t=1}^{T}  F_t (\Theta^t; \textbf{x}^t, \textbf{y}^t ) - \sum_{t=1}^{T} F_t (\Theta^*; \textbf{x}^t, \textbf{y}^t) \label{regret}
\end{align}
Here, $\Theta^* = \arg\min_\Theta \sum_{t=1}^{T} F_t (\Theta; \textbf{x}^t, \textbf{y}^t)$ represents the optimal fixed model selected in hindsight. Our objective is to minimize the learning regret, which equates to minimizing the cumulative loss. Importantly, if the learning regret grows sublinearly with respect to $T$, it indicates that the online learning algorithm can progressively reduce the training loss asymptotically.

The fixed optimal strategy in hindsight refers to a strategy determined by a centralized entity with full prior knowledge of all per-round loss functions. In our problem, achieving such an optimal strategy would require access to future information, including upcoming data collection for all rounds. However, this information is inherently unpredictable due to its randomness. As a result, the complete loss functions are also unknown at the outset and evolve dynamically over time. Therefore, regret just serves as a metric to quantify the performance gap between the proposed algorithm and the theoretical optimal strategy in our theoretical analysis. For experimental validation, we evaluate the learning performance of the proposed algorithm using metrics such as test loss and test accuracy.

\section{Denoising and Adaptive Online Vertical Federated Learning}
\begin{figure}[htp]
\vspace{-5pt}
\centering
\subfloat{\includegraphics[width=0.9\linewidth]{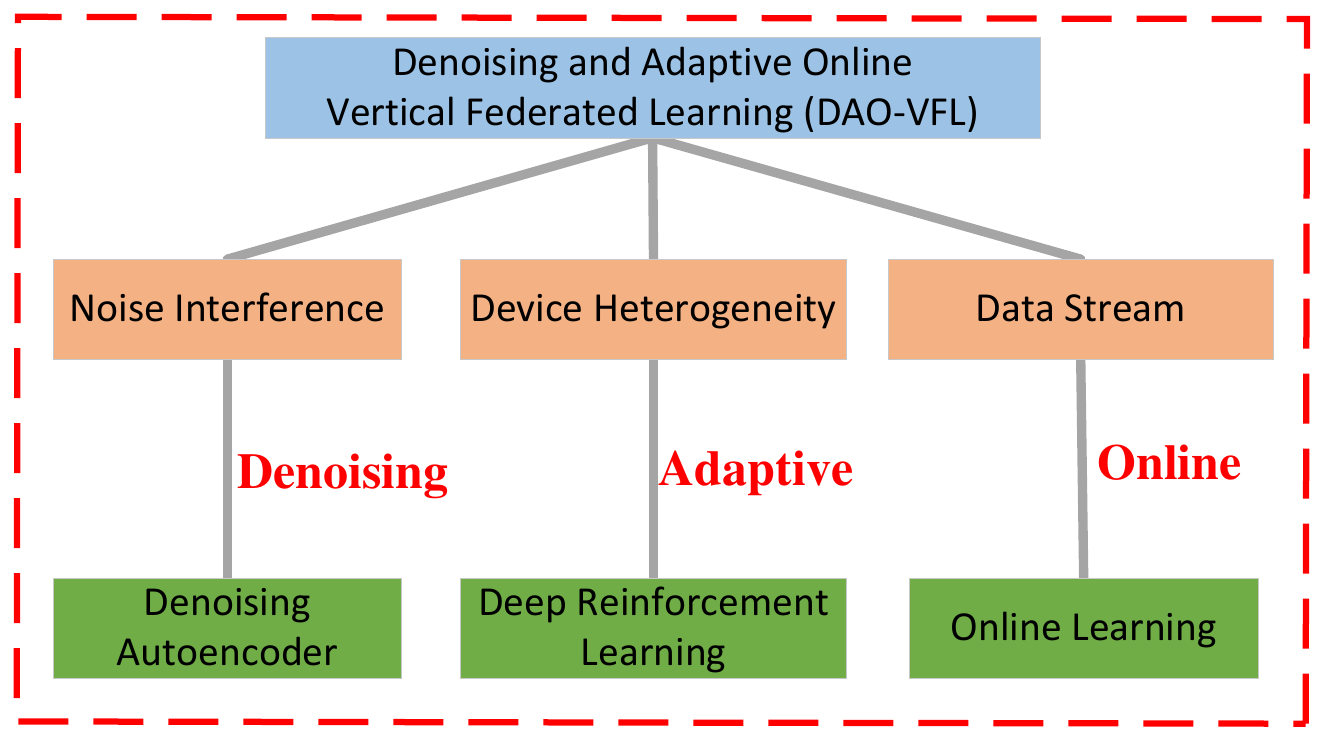}} 
\caption{Challenges and Solutions in DAO-VFL}  
\label{DAO-structure}
\vspace{-10pt}
\end{figure}

In this section, we introduce the details of the DAO-VFL algorithm. To provide an overview, we summarize the challenges faced by DAO-VFL and the corresponding solutions in Fig.\ref{DAO-structure}. Unlike traditional VFL approaches, the inherently complex environments of IIoT scenarios necessitate addressing three additional challenges. First, sensors in industrial wireless network environments are subject to noise interference \cite{cheffena2016industrial}, which can negatively impact training performance. Therefore, it is crucial to design effective and easily deployable noise reduction methods to mitigate noise effects. Second, the varying positions of sensors along the assembly line, coupled with differences in their computational capacities and channel conditions, add further complexity. Achieving synchronized updates within the learning system requires different sensors to perform varying numbers of local iterations. Developing an adaptive local iteration strategy that enhances overall training efficiency and performance is also a crucial aspect in this work, as it has not been adequately explored before. Third, sensor data in industrial assembly lines is collected in real-time. As a result, the algorithm must align with an online learning approach to effectively handle the dynamic dataset.

Next, we will outline the detailed process of the DAO-VFL algorithm, beginning with the introduction of some additional concepts. For convenience, a single time period in our framework is also defined as one global training round. During each global round $t \in \mathcal{T}$, both the sensors and the server execute a specified number of local training iterations, represented by the parameter $E_{t, k}$. This parameter is dynamically determined based on the real-time status of the sensors in the current global round. The approach for deciding the number of local iterations for each sensor will be discussed in the later section. Each local training iteration is indexed as $\tau = 0, 1, 2, ..., E_{t, k}$. Notably, the DAO-VFL algorithm is designed for an online synchronized scenario, even though sensors perform varying numbers of local training iterations. The detailed flow of the DAO-VFL algorithm is outlined in Algorithm \ref{alg: dao-vfl}. Subsequently, we provide an in-depth explanation of the key steps.

\begin{algorithm}
    \caption{DAO-VFL} \label{alg: dao-vfl} 
    \begin{algorithmic}[1]
        \State \textbf{Initialize}: The initial feature model $\theta_k^{t=0,\tau=0}$ for all sensors $k$ and the initial head model $\theta_0^{t=0,\tau=0}$ for server.
        \For {$t =0, 1, 2, ..., T - 1$}
            \For {Sensor $k = 1, 2, ..., K$ in parallel}
                \If {$\tau = 0$}
                    \State Collects new data samples $\textbf{x}_k^{t}$.
                    \State Gets the feature embedding ${h}_k ({\theta}_k^{t, 0}; \textbf{x}_k^t)$. 
                    \State Sends ${h}_k ({\theta}_k^{t,0}; \textbf{x}_k^t)$ to server.
                \EndIf    
            \EndFor
            \State Server collects noise feature embedding 
            \Statex \quad \quad   $\tilde{h}_k ({\theta}_k^{t,0}; \textbf{x}_k^t)$ from sensors.
            \State Server applies the denoising autoencoders 
            \Statex \quad \quad and get the denoised embedding  $\hat{h}_k ({\theta}_k^{t,0}; \textbf{x}_k^t)$.  
            \State Server collects model representation $\hat{\Phi}^{t,0}$.
            \State Server sends $\hat{\Phi}^{t,0}$ to all sensors.
            \For {$k =0, 1, 2, ..., K$ in parallel}
                \State Each sensor $k$ decide the $E_{t, k}$ via Algorithm \ref{alg: drl_tp}.
                \For {$\tau = 1, 2, ..., E_{t, k} -1$}
                    \State Gets $\hat{\Phi}^{t, \tau}_k \leftarrow \left \{ \hat{\Phi}^{t,0}_{-k}; {h}_k^t ({\theta}_k^{t, \tau})\right \}$.
                    \State Updates feature or head model ${\theta}_k^{t, \tau + 1}$. 
                \EndFor
                \State Inherit the model $ {\theta}_k^{t+1, 0} \leftarrow {\theta}_k^{t, E_{t, k}}$ for next round.
            \EndFor    
        \EndFor   
    \end{algorithmic}
\end{algorithm}


\subsubsection{Feature Embedding Extracting}
At the start of each global round $t$, each sensor $k$ in the assembly line incrementally collects the new training dataset, $\textbf{x}_k^{t}$ and $\textbf{y}^t$. It is important to note that the data samples $\textbf{x}_k^{t}$ are collected independently by each sensor in chunks, with each sensor capturing only one or several features. These data samples are then processed by each sensor's local feature model, $\theta^{t,0}_k$ to generate feature embeddings $h_k^t (\theta_k^{t,0})$. The initial feature model $\theta^{t,0}_k$ at current global round is inherited from the previous global round. Following this, each sensor uploads its feature embeddings to the server to further obtain the model representation.

\subsubsection{Feature Embedding Denoising}
Given the interference present in complex wireless network environments within the IIoT based assembly line scenario, feature embeddings transmitted through the wireless network are inevitably affected by noise. Suppose we define $\tilde{h}_k({\theta}_k^{t,0}; \textbf{x}_k^t)$ as the noisy feature embedding. To address this, it is essential to develop effective noise reduction techniques to obtain the denoised feature embedding, denoted as $\hat{h}_k({\theta}_k^{t,0}; \textbf{x}_k^t)$. To achieve this, we employ Denoising Autoencoders (DAE) for noise reduction on the server side. Given the server’s robust computational capabilities, this process does not impose a significant computational burden. DAE is a specialized type of neural network designed to learn robust feature representations by reconstructing input data from a corrupted version. Unlike traditional autoencoder, which focus on compressing and then reconstructing the original input, DAE is explicitly trained to remove noise from the input, thereby revealing the underlying structure of the data. 

The noise reduction process operates as follows: once the server receives the noisy feature embedding, it first uses an encoder to map the embedding from a high-dimensional space to a lower-dimensional latent space. The decoder then reconstructs the feature embedding, mapping it back to the original space. During the initial $T_{dl}$ global rounds, also referred to as the denoising learning period, it is assumed that the original feature embeddings from the sensors are available to the server.  Achieving optimal DAE performance requires close cooperation between the sensors and the server and is strongly influenced by the $T_{dl}$. Theoretically, a larger $T_{dl}$ allows for more effective noise reduction. Based on the network conditions, the server can employ either a single DAE or multiple DAEs to denoise the noisy feature embeddings received from the sensors. Then for the paired feature embedding $\left\{ \tilde{h}_k^t (\theta_k^{t,0}), h_k^t (\theta_k^{t,0}) \right\}$, the training target of DAE is:
\begin{align}
   \arg \min_{\theta_d} \mathbb{E}_{\left\{ \tilde{h}_k^t (\theta_k^{t,0}), h_k^t (\theta_k^{t,0}) \right\} } \left\{Ls \left (\Lambda_{\theta_d} (\tilde{h}_k^t (\theta_k^{t,0})), h_k^t (\theta_k^{t,0})  \right ) \right\}\label{dae_a}
\end{align}
where $\mathbb{E}$ denotes the expectation operator, $\Lambda_{\theta_d}(\cdot)$ represents the trainable DAE, $\theta_d$ refers to the weights of the DAE, and $Ls(\cdot)$ is the loss function used to evaluate the DAE's learning performance. Once processed through the trained DAE, the noisy feature embedding $\tilde{h}_k({\theta}_k^{t,0}; \textbf{x}_k^t)$ is transformed into the denoised feature embedding $\hat{h}_k({\theta}_k^{t,0}; \textbf{x}_k^t)$. In later sections, we will validate the effectiveness of noise reduction through both theoretical analysis and experimental results.

\begin{figure}[htp]
\vspace{-5pt}
\centering
\subfloat{\includegraphics[width=0.9\linewidth]{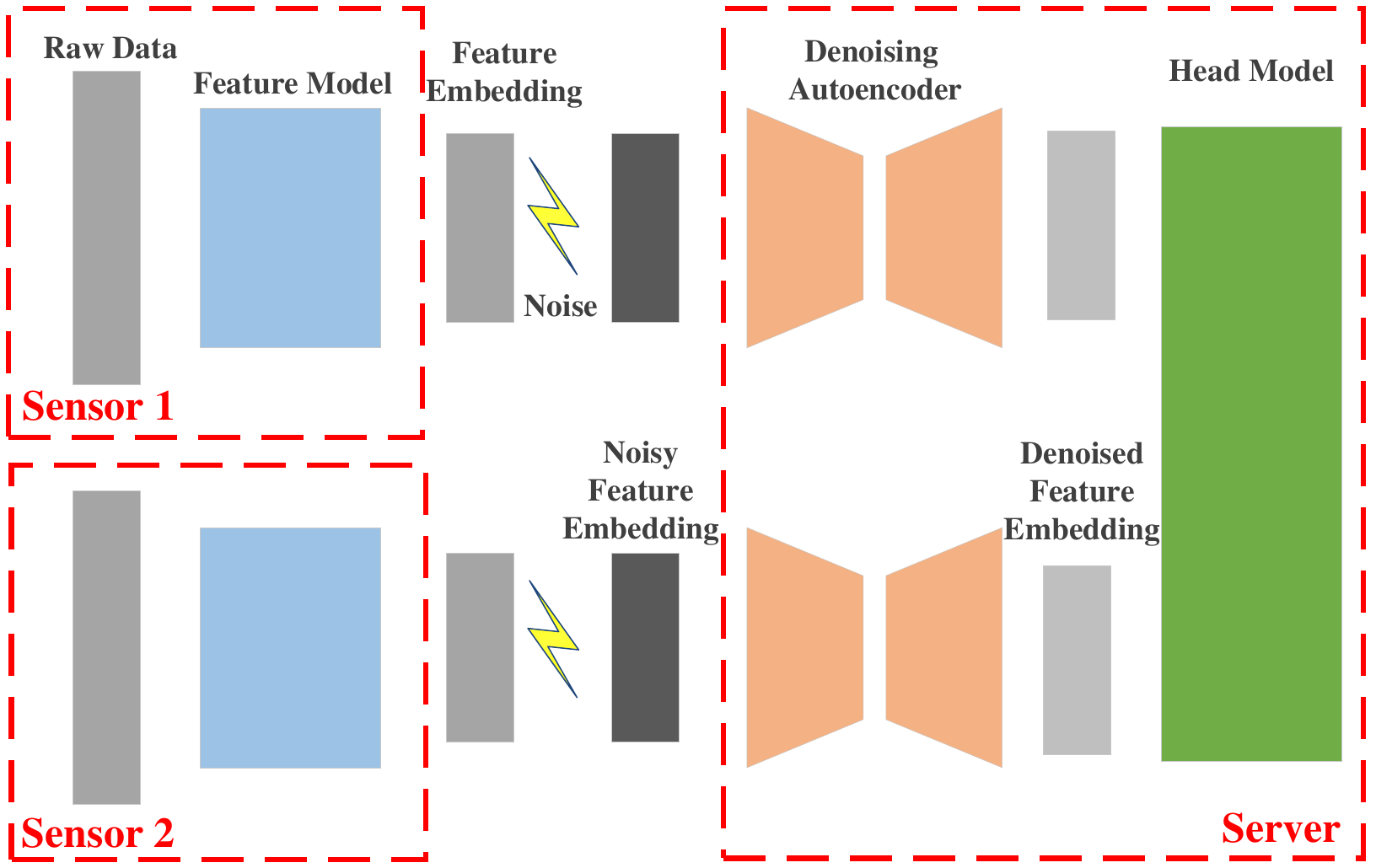}} 
\caption{The Architecture of DAO-VFL Noise Reduction}  
\label{DAO-NR}
\vspace{-10pt}
\end{figure}

\subsubsection{Model Representation Distributing}
After collecting all the denoised feature embedding, the server compiles the model representation $\hat{\Phi}^{t, 0}$, which includes the server head model and all denoised feature embeddings. The model representation is formally defined as follows:
\begin{align}
\hat{\Phi}^{t,0}\leftarrow \left \{ {\theta}_0^{{t,0}},   \hat{h}_1^t ({\theta}_1^{{t,0}}), \cdots, \hat{h}_k^t ({\theta}_k^{{t,0}}), \cdots,   \hat{h}_K^t ({\theta}_K^{{t,0}})\right \}  
\end{align}
The server then distributes the model representation $\hat{\Phi}^{t, 0}$ to all sensors. In this distribution process, the impact of noise is not considered. This is because servers can utilize directional antennas to focus transmission beams towards specific sensors, ensuring reliable communication.

\subsubsection{Feature and Head Model Updating}
Each sensor $k$ and the server employ the received model representation $\hat{\Phi}^{t, 0}$ to update their respective feature or head models over for $E_{t, k}$ iterations. Here, $E_{t, k}$ representing the adaptive local iteration decisions, varies across sensors. The updates follow this formula for all $\tau = 0, ..., E_{t, k} - 1$:
\begin{align}
    \theta_k^{t, \tau +1} = \theta_k^{t, \tau} - \eta \nabla_k F_t\left (\hat{\Phi}^{t,0}_{-k}, {h}_k^t(\theta_k^{t, \tau} ) \right ) \label{lu}
\end{align}
Here, $\hat{\Phi}^{t, 0}_{-k}$ is the collection of feature embeddings from all sensors and the head model from server, excluding sensor $k$. Notably, although the server and sensors may have different numbers of local iterations $E_{t, k}$, they operate within the same wall clock time, ensuring that the training process remains synchronized. Furthermore, since the training is based on a dynamic dataset rather than a subset of a static dataset, both the feature and head models are updated using online gradient descent (OGD) \cite{ying2008online} instead of traditional offline gradient descent methods. To enhance the understanding of the DAO-VFL process described earlier, we also present the DAO-VFL noise reduction model architecture in Fig.~\ref{DAO-NR}. 

\textbf{Remark 1}: Consistent with the scenarios outlined in previous VFL studies \cite{wang2023online, castiglia2022compressed}, we assume that all sensors and the server have access to label information. Furthermore, the IIoT based assembly line scenario is considered to operate in a low-risk environment, where label sharing between sensors and the server does not pose significant privacy concerns. 

\textbf{Remark 2}: To train the DAE, clean data are typically required. While this could theoretically be achieved by deploying additional costly redundant sensors or using dedicated communication lines \cite{zurawski2014industrial}, however, such approaches are not feasible for widespread, long-term deployment. An alternative method is to rely solely on noisy data for training. Notably, certain DAE-based noise reduction techniques, such as Noise2Noise \cite{lehtinen2018noise2noise}, Noise2void \cite{krull2019noise2void}, are specifically designed to function in scenarios where clean data is unavailable. These approaches can also be integrated into our algorithm. However, due to the complexity of the current algorithm, we do not explore these additional extensions in this work.
 
\section{Regret Analysis}
In this section, we provide a comprehensive regret analysis of the proposed DAO-VFL algorithm. To support our analysis, we begin by introducing several additional definitions. Specifically, we define $\hat{\textbf{G}}^{t}$ as the stacked partial derivatives that incorporate the effects of denoising at global round $t$:
\begin{align}
   \hat{\textbf{G}}^{t} := \left [  \sum_{\tau = 0}^{E_{t, 0} -1} \nabla_{0}  F_t (\hat{\Phi}^{t, \tau}_0), \dots,  \sum_{\tau = 0}^{E_{t, K} -1} \nabla_{K}  F_t (\hat{\Phi}^{t, \tau}_K) \right ]
\end{align}
where $\hat{\Phi}^{t, \tau}_k = (\hat{\Phi}^{t,0}_{-k}, {h}_k^t(\theta_k^{t, \tau} ))$. Using $\hat{\textbf{G}}^{t}$, we can define the updates to the global model during a global round $t$ with the following equation:
\begin{align}
\Theta^{t+1, 0} = \Theta^{t, 0} - \eta  \hat{\textbf{G}}^{t}
\end{align}
For comparison, we also define ${\textbf{G}}^{t}$ as the stacked partial derivatives at global round $t$ without accounting for the impact of noise:
\begin{align}
   {\textbf{G}}^{t} := \left [  \sum_{\tau = 0}^{E_{t, 0} -1} \nabla_{0}  F_t ({\Phi}^{t, \tau}_0), \dots,  \sum_{\tau = 0}^{E_{t, K} -1} \nabla_{K}  F_t ({\Phi}^{t, \tau}_K) \right ]
\end{align}
For the subsequent theoretical analysis, we consider the overall gradient and model as $D$-dimensional vector. We denote an arbitrary vector element $d \in [1, D]$  of the overall gradient as ${\textbf{G}}_{k, d}$, and the arbitrary vector element $d \in [1, D]$ of the overall model as $\Theta_{k, d}$.

Next, we will present the assumptions typically employed in the analysis of online convex optimization, as referenced in \cite{park2022fedqogd}. Some of these assumptions are defined at the vector element level and are specifically tailored to align with the requirements of our proof within the context of online learning in VFL scenarios \cite{wang2023online}.

\begin{assumption}
For any $(\textbf{x}^t; \textbf{y}^t)$, the loss function $F_t (\Theta; \textbf{x}^t; \textbf{y}^t)$ is convex with respect to $\Theta$ and differentiable.
\label{assm:cov}
\end{assumption}

\begin{assumption}
The loss function is $L$-Lipschitz continuous, the partial derivatives satisfies: $\left \| \nabla_{k}  F_t (\Theta)   \right \|^2 \leq L^2$.
\label{assm:bpd}
\end{assumption}

\begin{assumption}
The partial derivatives corresponding to a consistent loss function satisfy the following condition:
\begin{align}
\left \|  {\textbf{G}}_{k}^{t, \tau'} - {\textbf{G}}^{t, \tau}_{k} \right \| \leq \lambda \left \| \theta_k^{t, \tau'} - \theta_k^{t, \tau} \right \| \notag
\end{align}
\label{assm:gradient-change}
In the context of the online learning scenario, where the loss function changes over time, we use $t$ to indicate that gradients and models correspond to the same loss function. To differentiate their origins from various local iterations, we use $\tau'$ and $\tau$, respectively.
\end{assumption}

\begin{assumption}
The arbitrary vector element $d$ in the overall model $\Theta_{k, d}$ is bounded as follows: $\left | \Theta_{k, d} \right | \leq \rho $.
\label{assm:model-variant}
\end{assumption}

Next, we will provide the assumption for the overall gradient after noise reduction, comparing it to the original overall gradient.

\begin{assumption}
The arbitrary vector element $d$ in the overall gradient, adjusted through a denoising method, has a bounded range range of variation as:
$\left |\hat{\textbf{G}}^{t, \tau}_{k, d} - {\textbf{G}}^{t, \tau}_{k, d}  \right | \leq \beta_d $.
\label{assm:gradient-noise}
\end{assumption}

Assumption \ref{assm:cov} ensures the convexity of the function, allowing us to leverage the associated properties of convexity. Assumption \ref{assm:bpd} constrains the magnitude of the loss function's partial derivatives. Assumption \ref{assm:gradient-change} guarantees that the variation in partial derivatives remains within a specific range, consistent with the model's variation across different local iterations under a consistent loss function. Assumption \ref{assm:model-variant} defines the allowable range for any vector element within the overall model. Finally, Assumption \ref{assm:gradient-noise} bounds the impact of the application of noise reduction methods on the overall model relative to the original model. Based on the above assumptions, we can derive the main result, presented as Theorem \ref{thm:general}, as follows:
\begin{theorem}\label{thm:general}
Under Assumption 1-5, DAO-VFL with adaptive local iterations decisions $E_{t, k} \geq 1$, while considering the impact of denoising, achieves the following regret bound:
\begin{align}
 & {Reg}_T   = \sum_{t=1}^{T} \mathbb{E}_t \left [ F_t (\Theta^{t,0}; \textbf{x}^t; \textbf{y}^t)  \right ] - \sum_{t=1}^{T} F_t (\Theta^*; \textbf{x}^t; \textbf{y}^t) \notag \\
 & \leq \frac{  \left \| \Theta^{1, 0} - \Theta^* \right \|^2}{2 \eta E_{min} }  + \frac{\eta T D \beta^2_d  }{ E_{min}} + \frac{\eta T E_{max} L^2 K}{ E_{min}} \notag \\
  & + 2  D T \rho (\eta \lambda E_{max} L + \beta_d )  \\
  & E_{max} = \max_{t \in \mathcal{T}, k \in \mathcal{K}} E_{t, k}, \quad E_{min} = \min_{t \in \mathcal{T}, k \in \mathcal{K}} E_{t, k} \notag
\end{align}
\end{theorem}
\begin{proof}
The proof can be found in Appendix.
\end{proof}
Based on Theorem \ref{thm:general}, we derive the following findings: First by setting $\eta = \mathcal{O}(1/\sqrt{T})$, the DAO-VFL can achieve the regret bound of $\mathcal{O}(\sqrt{T} + T \beta_d)$ over $T$ time rounds. The regret bound is affected by the term $\mathcal{O}( T \beta_d)$, which reflects the effectiveness of the noise reduction method. Second, by incorporating Assumption \ref{assm:gradient-noise} and defining $\beta_n$ as the bounded range of variation between the noisy gradient and the original gradient, we can substitute $\beta_d$ with $\beta_n$ in the regret bound from Theorem \ref{thm:general} to derive the regret bound in the presence of noise. It becomes clear that the primary factors influencing the regret bound are the magnitudes of $\beta_n$ and $\beta_d$. If the noise reduction method is effective, it consistently leads to a tighter regret bound. This comparison will be demonstrated through experimental results in the later sections. Third, the tightness of the regret bound is also  influenced by $E_{max}$ and $E_{min}$. A smaller $E_{max}$ and a larger $E_{min}$ lead to a tighter bound. This highlights the importance of ensuring that all sensors are as similar as possible in terms of $E_{t, k}$ during each global round, aligning with the principle of fairness in the number of local iterations across all sensors. In the following, we will use simulations to further validate this finding.

\begin{figure}[h]
\centering
\vspace{-15pt}
\subfloat[Test Loss with C-MAPSS]{\includegraphics[width=0.49\linewidth]{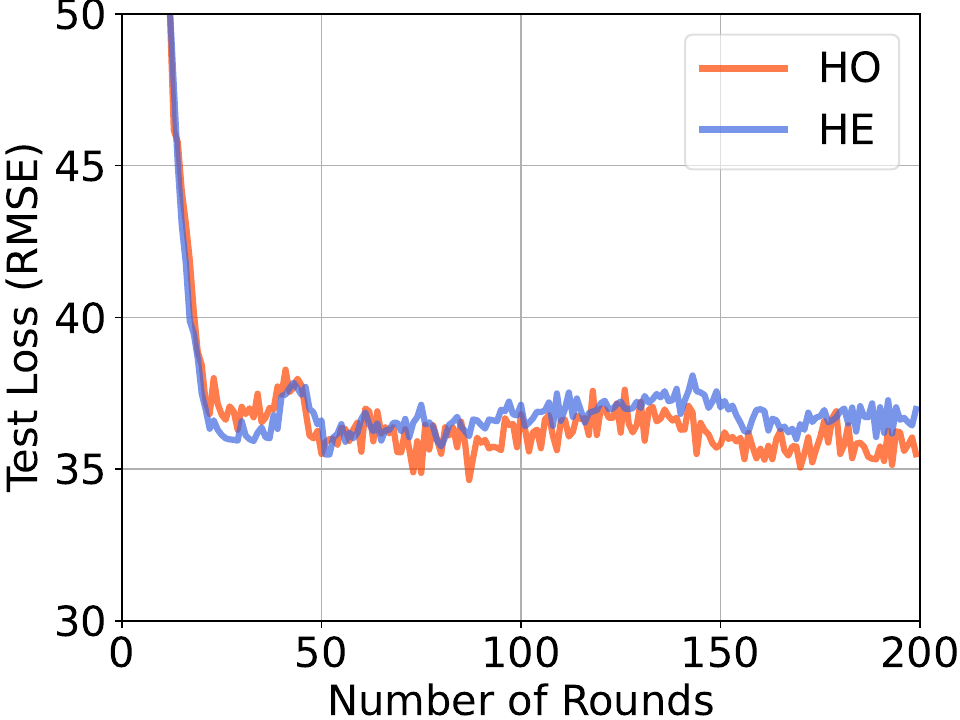}} 
\subfloat[Test Accuracy with CIFAR-10]{\includegraphics[width=0.49\linewidth]{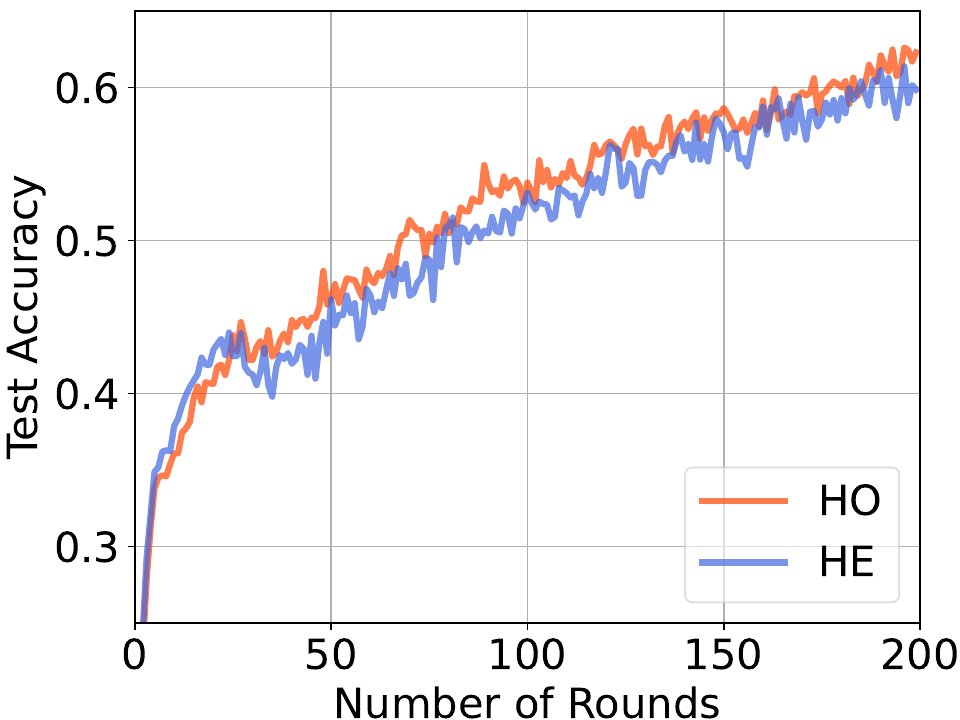}} 
\caption{Learning Performance of DAO-VFL with Different Local Iteration Decisions Patterns $(E_{t, k})$ .}   \label{deonvfl-ekt}
\vspace{-10pt}
\end{figure}

\textbf{Validation Study}: To validate the impact of fairness in the number of local iterations $E_{t, k}$ across all sensors, we conducted simulations using the C-MAPSS and CIFAR-10 datasets under different $E_{t, k}$ patterns. Further experimental details are provided in the experimental section. We tested two patterns: \textbf{Homogeneity} (HO), where all sensors had the same $E_{t, k}$ value in each global round, and \textbf{Heterogeneity} (HE), where sensors had significantly different $E_{t, k}$ values in each global round. To ensure a controlled comparison, the total local iterations, $  \sum_{k\in \mathcal{K}} E_{t, k}$ were kept constant across all global rounds for both patterns. The simulation results over 10 rounds are presented in Fig.~\ref{deonvfl-ekt}(a) for C-MAPSS dataset and Fig.~\ref{deonvfl-ekt}(b) for the CIFAR-10 dataset.

The experimental results reveal that maintaining similar $E_{t, k}$ values across sensors enhances learning performance, as reflected by higher test accuracy and lower test loss. This improvement is observed across both datasets. However, in real-world industrial scenarios, achieving identical $E_{t, k}$ values for all sensors at each global round can be challenging due to variations in sensors' computational and communication capabilities, particularly in synchronized training scenarios. In the following section, we will define the optimization problem specific to the IIoT-based assembly line scenario and utilize deep reinforcement learning to determine adaptive local iteration decisions for all sensors in each global round, taking sensor heterogeneity into account.

\section{Adaptive Local Iteration Decisions}
\begin{figure}[htp]
\vspace{-5pt}
\centering
\subfloat{\includegraphics[width=0.85\linewidth]{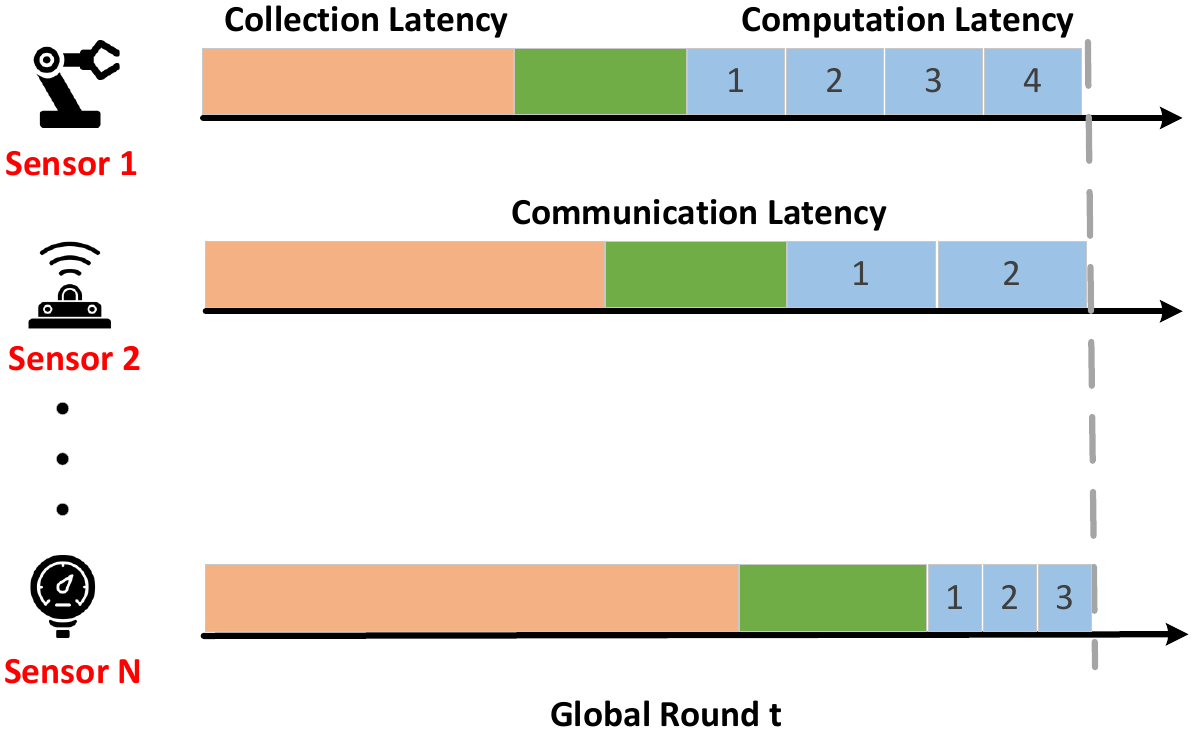}} 
\caption{Timeline of Sensors in One Global Round}  
\label{oleo}
\vspace{-10pt}
\end{figure}

In the IIoT-based assembly line scenario, sensors are positioned at various locations along the assembly line. Each sensor undergoes three primary phases within a single global training round: local data collection, feature embedding upload, and local feature model update. The detailed timeline of these phases is illustrated in Fig.~\ref{oleo}. In this problem, we disregard the time needed for the server to broadcast the model representation and omit the time spent on server-side noise reduction, as these factors are not directly relevant to the optimization problem and is not influenced by the sensors' capacities. Next, we will provide definitions for each of the three distinct phases in detail and present a comprehensive formulation of the optimization problem.

\subsubsection{Collection Latency}
Given that our scenario is based on online learning, where training is conducted using real-time collected data rather than pre-selected batches from a static dataset, data collection latency becomes a crucial factor. Additionally, in the context of an assembly line, we consider the intervals between the sequential data collection times of each sensor. To simulate this sequential process characteristic of an assembly line, we define the \textbf{collection latency} as follows:
\begin{align}
\Upsilon_{t, k}^{co} =  \mu  k +  \mu_0
\end{align}
For simplicity in the analysis, we assume that sensors collect data sequentially based on their index, such that sensors with smaller indices collect data earlier. 

\subsubsection{Communication Latency}
In this scenario, sensors share bandwidth during the upload process, as they operate within a wireless network and transmit their feature embeddings to the server via this wireless network. The \textbf{communication latency} is equal to:
\begin{align}
\Upsilon_{t, k}^{cm} =  \frac{W_{fe}}{r_{t, k}}
\end{align}
where $W_{fe}$ represents the number of weights in the feature embedding. We assume that the network bandwidth is equally distributed among all sensors. The transmission rate between sensor $k$ and the server at round $t$, denoted by $r_{t, k}$, as:
\begin{align}
r_{t, k} =  \frac{B}{K} \log_2 \left ( 1+ \frac{g_{t, k} p_k}{\sigma^2}\right )
\end{align}
where $B$ is the total bandwidth, $g_{t, k}$ represents the channel gain between sensor $k$ and the server at global round $t$, $p_k$ denotes the transmission power of sensor $k$, and $\sigma^2$ is the noise power.

\subsubsection{Computation Latency}
In each global round, the sensor performs at least one local update, with the number of updates depending on its specific conditions. The \textbf{computation latency} can be computed as follows:
\begin{align}
\Upsilon_{t, k}^{cp} =  \frac{E_{t, k} C W_{lc}}{f_{t, k}}
\end{align}
Here, $C$ represents the number of CPU cycles required for sensor to update a single model weight, and $W_{lc}$ denotes the number of weights in the local model. The CPU frequency of sensor $k$ during computation at round $t$ equals $f_{t, k}$, which varies across sensors and ranges from $f_k^{\min}$ to $f_k^{\max}$.

Based on the definitions of latency provided above, the total latency for each global round $t$ can be expressed as:
\begin{align}
\Upsilon_{t} = \max_{1 \leq k \leq K} \left (\Upsilon_{t, k}^{co} + \Upsilon_{t, k}^{cm} + \Upsilon_{t, k}^{cp} \right )
\end{align}

To ensure fairness and reduce disparities in local iteration decisions among sensors during each global round, as indicated by the theoretical analysis and validation study from previous section, we define the local iteration disparity for global round $t$ as:
\begin{align}
H_{t} & = \sum_{k = 1}^K \left|  E_{t, k} - \bar{E}_{t}  \right|  =  \frac{1}{K} \sum_{k = 1}^K | K E_{t, k} - \sum_{k = 1}^K E_{t, k} | 
\end{align}
Defining $H_t$ in this manner minimizes the disparity in local iterations among sensors.

Taking into account the definitions of total latency and the insights on disparity, it is clear that the optimization problem must strike a balance between test accuracy and these two additional components. Based on this, we define the optimization problem $\textbf{P1}$ as follows:
\begin{align}
    \textbf{P1}: & \max_{\mathcal{E}} \mathbb{E} \left [ \alpha_1 \sum_{t\in T}Acc(t)- \alpha_2 \sum_{t\in T} \Upsilon_{t}-\alpha_3\sum_{t\in T} H_t \right ]
\end{align}
where $\mathcal{E} = \left[ E_{t, k} \right] $ represents a $T \times K$ matrix for each sensor's local iterations decision with $E_{t, k} \in [1, E_{max}]$, here $E_{max}$ denotes the upper limit of local iterations. The parameters $\alpha_1$, $\alpha_2$, and $\alpha_3$ control the relative importance of each objective, as determined by the designers. Our objective is to solve the optimization problem $\textbf{P1}$, Given the large number of parameters and their complex interdependencies, which make direct solutions challenging. We utilize  Deep Reinforcement Learning (DRL) techniques \cite{arulkumaran2017deep}, including the actor-critic method \cite{konda1999actor} and the Proximal Policy Optimization (PPO) algorithm \cite{schulman2017proximal}, to solve the optimization problem $\textbf{P1}$ and derive the adaptive local iteration decisions for each sensor.

The objective of the DRL agent is to find the best policy mapping a state to an action that maximizes the expected reward. In the following, we will provide a detailed explanation of the state space, action space, reward and training methodology relevant to the proposed DRL problem. 

\textbf{State}: 
The state consists of the information uploaded by each sensor prior to the local feature model update phase of each global round. The state in the DRL framework is represented as vectors: the data collection latency vector for round $t$, $\hat{\Upsilon}_{t}^{co} = (\Upsilon_{t, 1}^{co}, \dots \Upsilon_{t, K}^{co})$, the communication latency vector for round $t$, $\hat{\Upsilon}_{t}^{cm} = (\Upsilon_{t, 1}^{cm}, \Upsilon_{t, K}^{co})$, and CPU frequency vector of sensors for round $t$, $\hat{f}_{t} = ({f}_{t, 1}, \dots, {f}_{t, K})$, all uploaded by the sensors. The state vector $\textbf{S}_{t}$ for the DRL framework at round $t$ is defined as a vector with the following four components:

\begin{align}
\textbf{S}_{t} = [\hat{\Upsilon}_{t}^{co}, \hat{\Upsilon}_{t}^{cm}, \hat{f}_{t}, t ]
\end{align}
To expedite the training of DRL, we normalize each element in the state vector to ensure they are on the same scale.

\textbf{Action}: At global round $t$, the DRL agent generates a local iteration decision for all sensors as the action based on the state collected and uploaded to the server, This action is defined as:
\begin{align}
\textbf{A}_{t} = [E_{t, 1}, \cdots, E_{t, k}, \cdots, E_{t, K}], \quad E_{t, k} \in [1, E_{max}]
\end{align}
where the action space is discrete, and ${E}_{max}$ denotes the empirically predefined global upper limit for local iterations of all sensors. 

\textbf{Reward}: To optimize the FL performance outlined in \textbf{P1}, the reward function should capture the changes in learning performance, total latency, and local iteration disparity. Learning performance is measured after sensors execute the specified actions to update the model and subsequently evaluate it on the test datasets. Total latency and local iteration disparity are computed directly by the server based on the selected actions. Then the reward $\textbf{R}_t$ at global training round $t$ is defined as:
\begin{align}
\textbf{R}_t = \alpha_1 Acc(t)- \alpha_2  \Upsilon_{t} - \alpha_3 H_t
\end{align}
The reward can be derived from the $\textbf{P1}$ problem by decomposing it into subproblems for each global round, allowing the reward for each specific time slot to be computed directly.

\begin{algorithm}
    \caption{The DRL Agent Training Process} \label{alg: drl_tp} 
    \begin{algorithmic}[1]
        \State Randomly initialize actor network $\pi(\cdot)$ and critic network $V(\cdot)$ with weight $\theta_{a}$ and $\theta_{v}$.
        \State Initialize experience replay buffer $\mathcal{D}$.
        \For {$t =0, 1, 2, ..., T_{ag} - 1$}
            \State Each sensor records the information $(\Upsilon_{t, k}^{co}, \Upsilon_{t, k}^{co}, {f}_{t, k})$.
            \State Each sensor uploads its information to server.
            \State The server integrate the state $\textbf{S}_{t}$.
            \State Get action $\textbf{A}_{t}$ by feeding $\textbf{S}_{t}$ into the \hangindent=2em \hangafter=1 actor network.
            \State Each sensor performs a number of local updates based \Statex \quad \quad on the action $\textbf{A}_{t}$.
            \State Infer on test dataset to obtain test accuracy.
            \State The server calculate the reward $\textbf{R}_t$.
            \State Update the state of FL from $\textbf{S}_{t}$ to $\textbf{S}_{t+1}$.
            \State Store transition sample $(\textbf{S}_{t}, \textbf{A}_{t}, \textbf{R}_t, \textbf{S}_{t+1})$ into $\mathcal{D}$.
            \For {$m =0, 1, 2, ..., M$}
                \State The server update the actor network  $\theta_{a}$ using PPO.
                \State The server update the critic network $\theta_{v}$  by 
                \Statex \quad \quad \quad maximizing the reward via Eq.\ref{cn_loss}.
            \EndFor
        \EndFor
    \end{algorithmic}
\end{algorithm}

\textbf{Training Methodology}: 
PPO provides a balanced approach by combining ease of implementation, sample efficiency, and straightforward tuning. It is designed to compute updates that minimize the objective function while constraining deviations from the previous policy. Additionally, PPO is well-suited for scenarios involving discrete action spaces. Consequently, in this work, the DRL agent's actor network update process employs the PPO algorithm. 

The detailed training process of the DRL agent is outlined in Algorithm \ref{alg: drl_tp}. At the start of the DRL agent’s training process, the parameters of both the actor and critic networks are randomly initialized, and the experience replay buffer is set up. During each agent training round $t =0, 1, 2, ..., T_{ag} - 1$, sensor $k$ uploads the observation to the server, including the data collection latency, the communication latency and CPU frequency. The server consolidates these observations into a unified state and determines the local iteration decisions as action for all sensors by inputting the state into the actor network $\pi(\cdot)$. Each sensor subsequently executes the specified number of local iterations based on the action it receives. Once the local updates are completed, the server evaluates the test accuracy and calculates the reward, factoring in test accuracy, overall latency, and local iterations disparity. The training then transitions to the next state, $\textbf{S}_{t+1}$, while the experience from round $t$ is stored in the experience replay buffer. Then server subsequently updates the DRL agent using the experiences stored in the replay buffer, performing $M$ times. During this process, the actor network $\pi(\cdot)$ is updated by the PPO algorithm and the critic network $V(\cdot)$ is updated with the following gain function:
\begin{align}
\max_{\theta_{v}} \frac{1}{\left| \mathcal{D} \right|}\sum_{t=1}^{\left| \mathcal{D} \right|}\left [ \textbf{R}_t + \gamma V(\textbf{S}_{t+1};\theta_{v}) - V(\textbf{S}_{t};\theta_{v})  \right ]^2 \label{cn_loss}
\end{align}
After the DRL agent has been trained for $T_{ag}$ rounds, the server retains the actor network. During the subsequent DAO-VFL learning process, as outlined in Algorithm \ref{alg: dao-vfl}, the state is input into the actor network to generate the output action. This action is then used to guide the local iteration decisions of the sensors in each global round, enabling adaptive local iteration in DAO-VFL.

\section{Experiments}
In this section, we present the experimental evaluation of the DAO-VFL algorithm. The experiments were conducted on an Ubuntu 18.04 machine equipped with an Intel Core i7-10700KF 3.8GHz CPU and a GeForce RTX 3070 GPU. The model training module was built upon PyTorch. The detailed experimental settings are outlined below.
\subsection{Datasets}
To simulate DAO-VFL in IIoT assembly line scenarios, we utilize the CIFAR-10 dataset, a widely used benchmark, alongside the real-world IIoT sensor-based dataset, C-MAPSS. Detailed descriptions of both datasets are provided below.

\textbf{CIFAR-10}: The CIFAR-10 dataset is widely used for image classification tasks, containing a total of 60,000 32x32 color images in 10 distinct classes. In the training setup, 4 sensors are involved, with each sensor handling a specific quadrant of every image. At each global round, the trained models are evaluated on the test dataset to measure the current test accuracy.

\textbf{C-MAPSS}: The C-MAPSS (Commercial Modular Aero-Propulsion System Simulation) dataset \cite{saxena2008damage}, created by NASA, is extensively utilized in research on Remaining Useful Life (RUL) prediction, particularly in the field of aerospace engineering for prognostics. This dataset models the degradation processes in aircraft turbofan engines under a range of operational and fault conditions. It contains four subsets, each varying in the number of operating and fault conditions, with each subset divided into training and test sets. For our experiments, we use the FD002 subset, consisting of 50,119 data samples for training and 30,365 data samples for testing. Each row in the dataset provides a snapshot from a single operating cycle and contains 27 columns: the first column indicates engine ID, the second the current operational cycle number, columns 3-5 represent three operational settings influencing engine performance, columns 6-26 capture readings from 21 sensors, and the 27th column shows the actual RUL. In this setup, we assume the data is collected among 2 sensors.

Each time series begins with the engine operating under normal conditions, with a fault developing at an unknown point in time. In the training set, this fault progresses in severity until it results in system failure. In the test set, data is available up to a point shortly before failure. The objective is to predict the number of operational cycles remaining before failure in the test data. The performance of the RUL estimation model is evaluated using Root Mean Square Error (RMSE), a widely used metric for assessing RUL estimation.

\subsection{Online Data Generation} In the experiment, operating within an online learning scenario requires the training dataset to be dynamic, with data collected at the start of each global round. To ensure sufficient data samples for good training performance, we collect the initial dataset at the beginning of the training process. Considering the differences in dataset types and sizes, distinct online data generation methods are employed for the CIFAR-10 and C-MAPSS datasets. Details for each case are provided below.

\textbf{CIFAR-10}: For the CIFAR-10 dataset, as the data samples are independent and lack time-series correlation, we adopt a fixed-size training dataset approach for each global round. In this approach, for every new data sample added, an equal number of older data samples are removed. The initial training dataset contains 5,000 samples. Starting from the first global round, each sensor collects 200 new data samples corresponding to its partial features and removes 200 old data samples in each global round. This ensures that training dataset is consistently updated every global round with the same amount of data samples. Similarly, the test dataset is updated in the same manner every global round.

\textbf{C-MAPSS}: For the C-MAPSS dataset, where data samples are interconnected due to their representation of the RUL of a aircraft turbofan engines over time, we adopt an incremental training dataset approach. In this case, the training dataset gradually accumulates new data samples until it encompasses the entire original training dataset. The initial training dataset contains 1,000 samples. After the first global round, each sensor collects 100 new data samples for the corresponding partial features in each subsequent global round. For the test dataset, the full set of samples from the original test dataset is used in this case.

\subsection{Model Details} Here, we will individually present all the models used in our experiments, including the feature model, head model, denoising autoencoder model, and the Actor and Critic network models for DRL, separately for the two datasets. 

\textbf{CIFAR-10}: The feature model consists of 13 convolutional layers followed by pooling layers. It concludes with fully connected layers, and outputs a feature embedding of size 4096. The head model is a single fully connected layer that takes the concatenated feature embeddings from all sensors (4096 × 4) and maps them to the final 10 classes. The feature and head model employs the OGD optimizer with a learning rate of 0.01. The denoising autoencoder model includes an encoder with three fully connected layers, taking an input of size 4096 and producing an output of size 512. It also features a decoder with three fully connected layers, taking an input of size 512 and generating an output of size 4096. The DAE model employs the Adam optimizer with a learning rate of 0.01. The Actor network consists of three fully connected layers, each with 64 neurons. The input size of the Actor network corresponds to the state size, while the output size matches the action size. Similarly, the Critic network comprises three fully connected layers with 64 neurons each. It takes the state as input and outputs a single state-value, estimating the expected return from a given state under the current policy. The Actor model uses the Adam optimizer with a learning rate of 0.0001, while the Critic model uses the Adam optimizer with a learning rate of 0.001.

\textbf{C-MAPSS}: The feature model consists of two convolutional layers with 8 and 14 channels, respectively, followed by a flattening layer that outputs a feature embedding of size 28 for each sensor. The head model is a single fully connected layer that processes the concatenated feature embeddings from all sensors (28 × 2) and maps them to the final 131 classes for the classification task. The feature and head model employs the OGD optimizer with a learning rate of 0.01 The denoising autoencoder model includes an encoder with three fully connected layers, taking an input of size 28 and producing an output of size 3. It also features a decoder with three fully connected layers, which takes an input of size 3 and reconstructs an output of size 28. The DAE model employs the Adam optimizer with a learning rate of 0.01. The Actor and Critic networks share a similar structure to those described earlier, with the only difference being the size of the action space, which varies depending on the number of sensors involved. The Actor model uses the Adam optimizer with a learning rate of 0.0001, while the Critic model uses the Adam optimizer with a learning rate of 0.001.

\subsection{Environment Details} 
Here, we address the environment setup considerations for the adaptive local iteration decisions part. The setup focuses on designing collection latency, communication latency, and computation latency. To ensure a balanced contribution from each component and prevent any single factor from dominating due to significantly higher values, we adjust their values to be of the same order of magnitude. For collection latency, we set $\mu_0 = 2$, and $\mu$ is assigned a random value between $[2, 4]$. For communication latency, we set the number of weights $W_{fe} = 1 \times 10^5$, total bandwidth $B = 1 \times 10^7$HZ, transmission power $p_k=1$W, noise power $\sigma^2 = 5\times 10^{-2}$W, and the channel gain is a random value between $[1 \times 10^{-4}, 1 \times 10^{-5}]$, varying across sensors and rounds. For computation latency, 
we set the number of CPU cycles for a single model weight $C =1000$ and the number of weights in the local model to $W_{lc} =5 \times 10^{5}$, the CPU frequency of sensor is selected from two categories: high performance sensors are chosen randomly from $[2 \times 10^{7}, 4 \times 10^{7}]$, and ow-performance sensors are chosen from $[1 \times 10^{7}, 3\times 10^{7}]$ in each global round. 

To summarize, the setup reflects the sensor heterogeneity and communication environment variations, specifically in terms of channel gain and CPU frequency, which align with the challenges present in IIoT-based assembly lines.

\subsection{Benchmarks} In the experiment, the following benchmarks are employed for performance comparison. Since the two key features of the DAO-VFL algorithm: noise reduction and adaptive local iteration decisions are essentially independent and do not influence each other, a control variable approach is employed to design independent benchmarks for comparison. We begin by presenting the benchmarks related to noise reduction part. 

\textbf{Noise Excluded (NE)}. In this approach, the communication process assumes no noise impact, enabling the server to receive noise-free feature embeddings from the sensors.

\textbf{Noise Included (NI)}. In this approach, communication noise is not addressed, resulting in the server receiving noisy feature embeddings from the sensors, which are then used in subsequent processes. Here, we consider using a uniform scalar quantizer to model the noise added to the feature embeddings, where the noise level is influenced by the quantization level. A smaller quantization level corresponds to higher noise.

Next, we will present the benchmarks related to adaptive local iteration decisions part. 

\textbf{Homogeneity (HO)}. In this approach, all sensors are trained using the same maximum number of fixed local iterations $E_{t, k} = E_{max}$. This scenario completely disregards the impact of total latency.

\textbf{Heterogeneity (HE)}. In this approach, sensors are trained with significantly different predetermined numbers of fixed local iterations. Specifically, one sensor is trained with $E_{max}$, while all other sensors perform the minimum local iterations, set as $E_{t, k} = 1$.

\subsection{Simulation Results}
Next, we present the experimental results of DAO-VFL, starting with an analysis of the impact of noise reduction on learning performance. To ensure a controlled comparison, we initially disregard the effect of local iteration decisions and assume that both our proposed method and the benchmarks follow identical local iteration decisions. All simulation results are averaged over 10 random runs.

\begin{figure}[h]
\centering
\vspace{-15pt}
\subfloat[Test Loss with C-MAPSS ]{\includegraphics[width=0.49\linewidth]{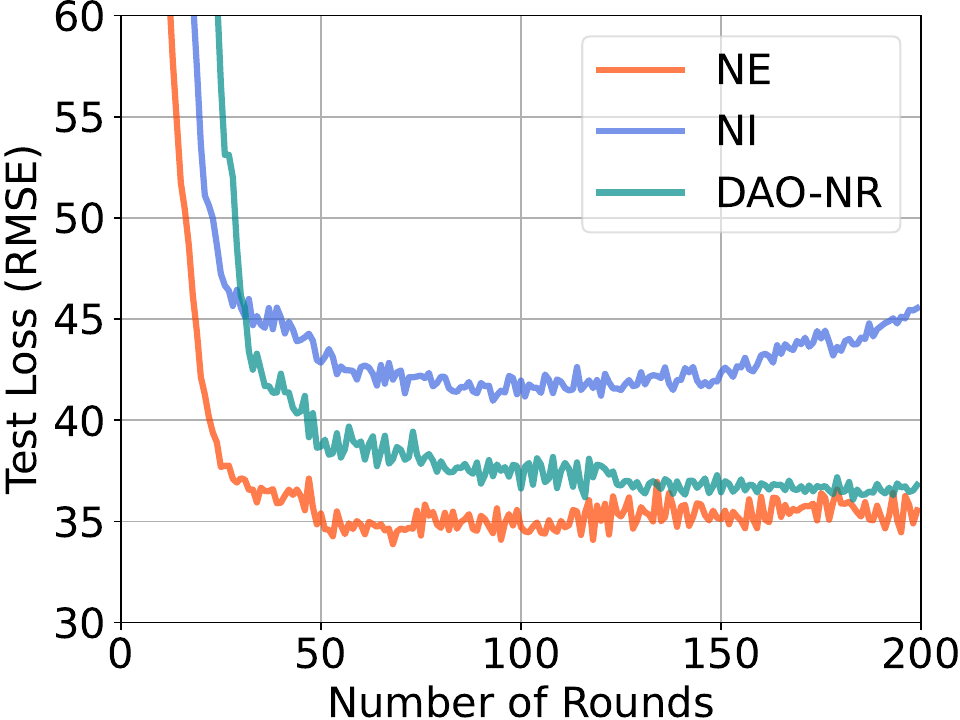}} 
\subfloat[Test Accuracy with CIFAR-10 ]{\includegraphics[width=0.49\linewidth]{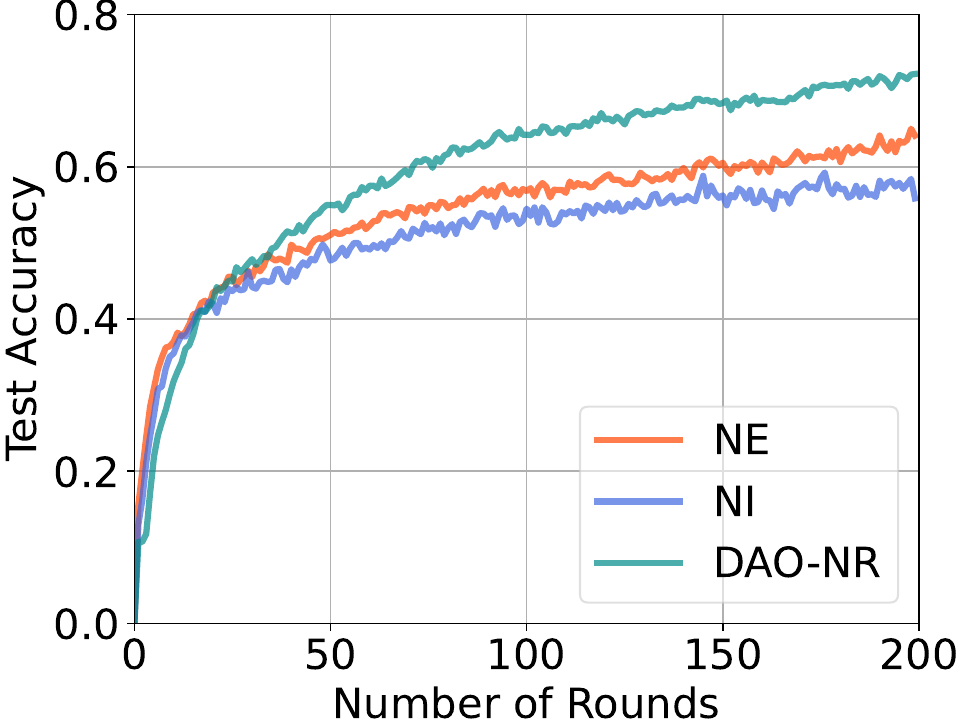}} 
\caption{Performance Comparison of DAO-VFL and Benchmarks Considering Noise Reduction Effects.}   \label{deonvfl-pc}
\vspace{-10pt}
\end{figure}

\textbf{Performance Comparison (Noise Reduction)}. We first evaluate the learning performance between proposed algorithm and benchmarks with $E_{t, k} = 2$ and $T_{dl} = 40$. Given the aforementioned simulation setup, the performance comparison between DAO-VFL and benchmarks is shown in Fig.~\ref{deonvfl-pc}(a) and Fig.~\ref{deonvfl-pc}(b) based on C-MAPSS dataset and the CIFAR-10 dataset, respectively. From both figures, several key observations can be drawn. First, we observe that noise significantly impacts learning performance across all datasets. This effect is particularly pronounced in the case of the C-MAPSS dataset, where failing to address noise can even hinder convergence. Second, we observe that the learning performance of the DAO-NR approach varies across different datasets. For the C-MAPSS dataset, the DAO-NR approach effectively mitigates the impact of noise, achieving learning performance comparable to that of the NE. For the CIFAR-10 dataset, the DAO-NR approach not only eliminates the effect of noise but also further enhances learning performance. This improvement may be attributed to the regularization effect of the DAE and its robust feature learning capabilities. In summary, the DAO-NR approach for noise reduction is both effective and essential for mitigating noise during online VFL training process.

\begin{figure}[h]
\centering
\vspace{-15pt}
\subfloat[Test Loss with C-MAPSS ]{\includegraphics[width=0.49\linewidth]{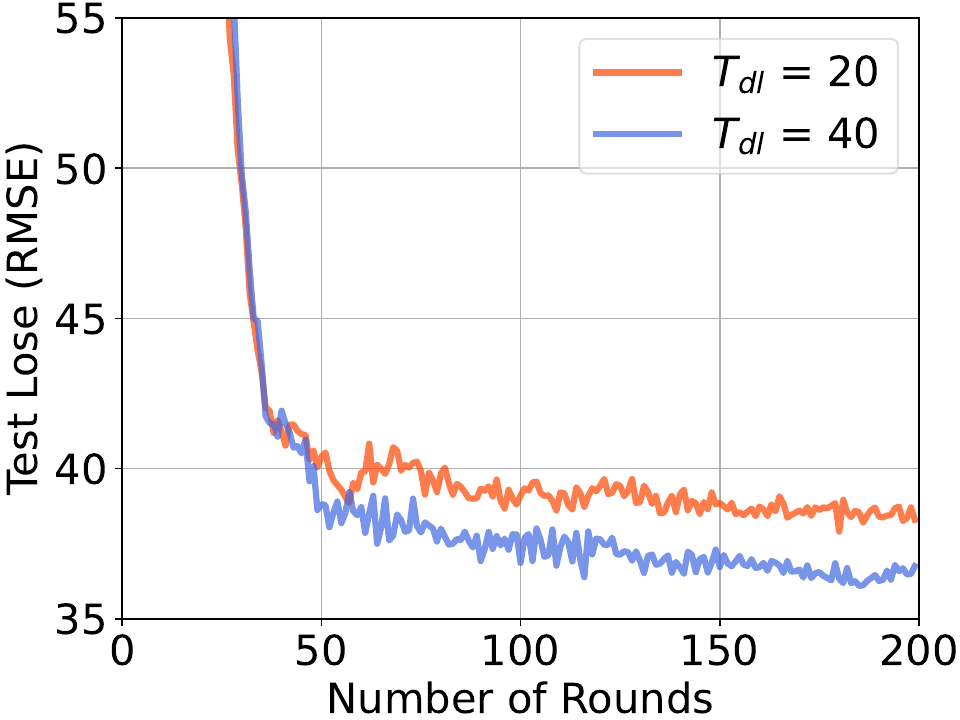}} 
\subfloat[Test Accuracy with CIFAR-10 ]{\includegraphics[width=0.49\linewidth]{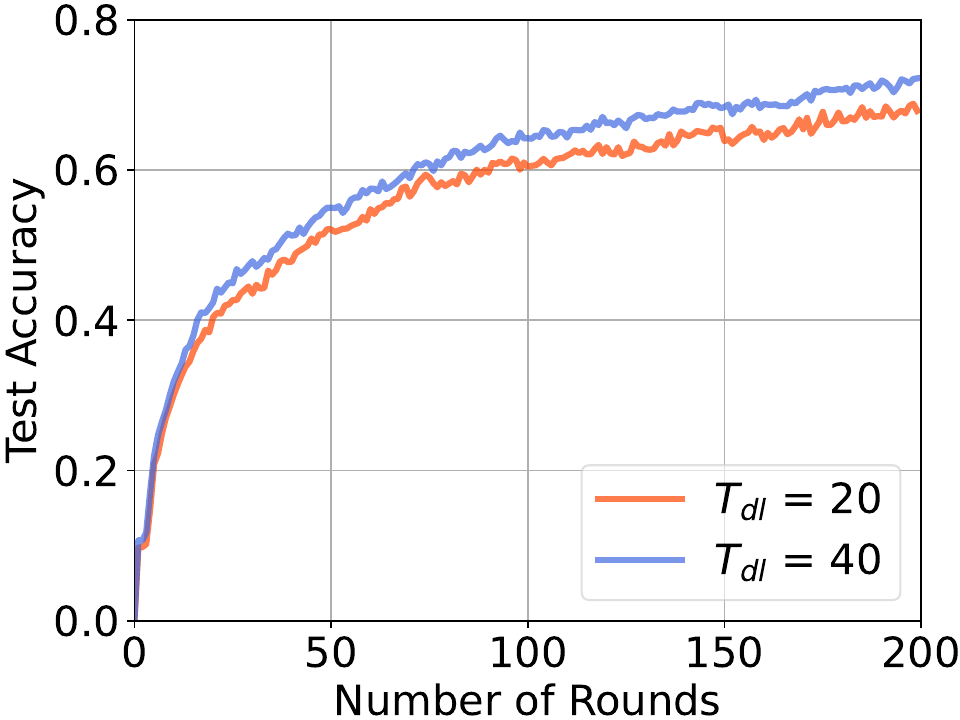}} 
\caption{Performance Comparison of DAO-VFL and Benchmarks with Different Denoising Learning Period $T_{dl}$.}   \label{deonvfl-dl}
\vspace{-10pt}
\end{figure}

\textbf{Impact of Denoising Learning Period $T_{dl}$}. In this section, we evaluate the impact of denoising learning period $T_{dl} \in \left [20, 40 \right ]$ on the learning performance. Fig.~\ref{deonvfl-dl}(a) and Fig.~\ref{deonvfl-dl}(b) depict the relationship between test loss/accuracy and the denoising learning period $T_{dl}$ for the C-MAPSS and CIFAR-10 datasets, respectively. The results indicate that extending the denoising learning period $T_{dl}$ improves learning performance in both cases. Notably, the improvement is more pronounced for the C-MAPSS dataset than for the CIFAR-10 dataset. This difference may be attributed to the incremental dataset approach used for the C-MAPSS dataset, where the increasing data volume significantly enhances the training of the DAE. In contrast, the CIFAR-10 dataset employs a fixed-size dataset approach, which limits the amount of data available for training the DAE.

Next, we will examine the impact of local iteration decisions on learning performance. To ensure consistency, we control variables so that the effect of noise reduction remains the same for both DAO-VFL and the benchmarks. We begin by comparing the learning performance between the adaptive local iteration decisions, derived using PPO, with the previously introduced benchmarks.

\begin{figure}[htbp]
\centering
\vspace{-15pt}
\subfloat[Test Loss with C-MAPSS]{\includegraphics[width=0.49\linewidth]{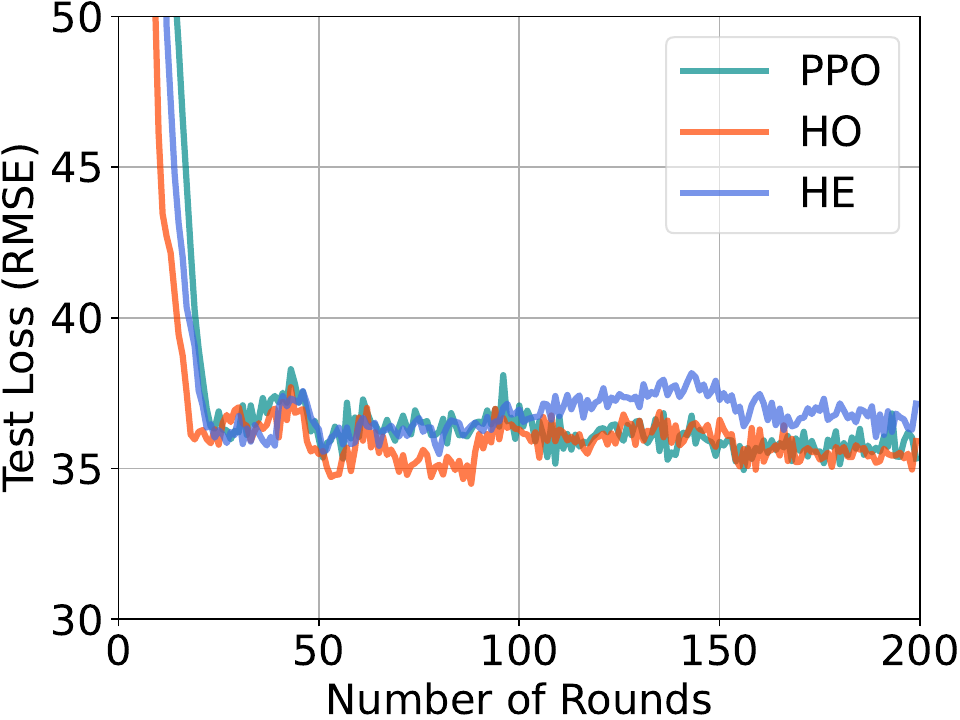}} 
\subfloat[Test Accuracy with CIFAR-10]{\includegraphics[width=0.49\linewidth]{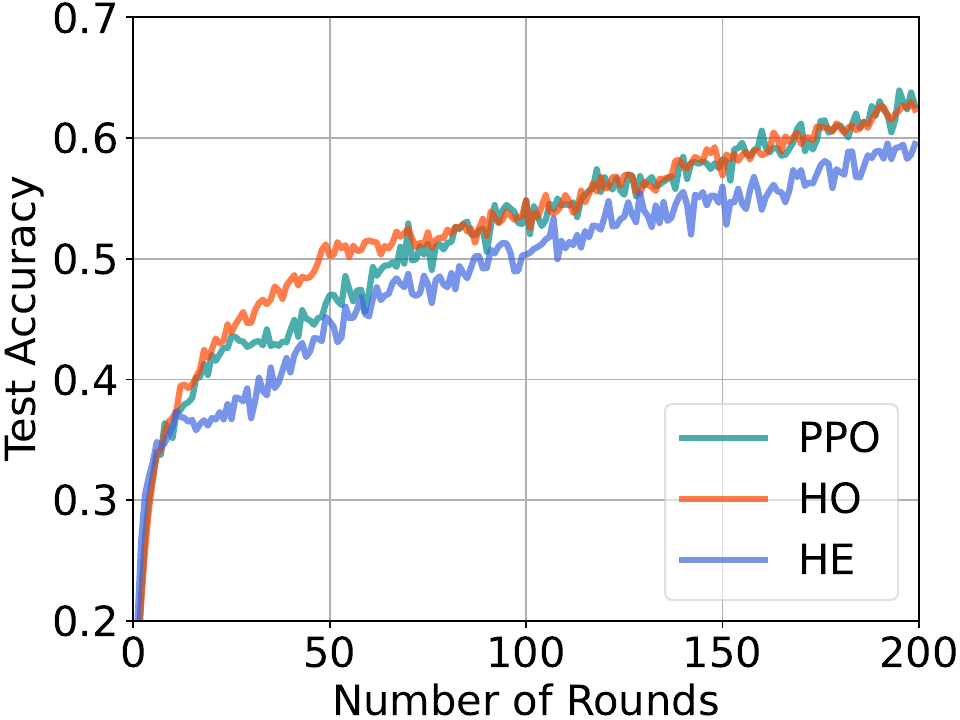}} 
\caption{Performance Comparison of DAO-VFL and Benchmarks Considering Adaptive Local Iteration Decision.}   \label{deonvfl-pclid}
\vspace{-10pt}
\end{figure}

\textbf{Performance Comparison (Adaptive Iterations)}. In this section, we evaluate the learning performance of DAO-PPO compared with the benchmarks HO and HE. Using the simulation setup described earlier, we present the performance comparison between DAO-PPO and the benchmarks, considering the local iteration decisions, is illustrated in Fig.~\ref{deonvfl-pclid}(a) and Fig.~\ref{deonvfl-pclid}(b) based on C-MAPSS dataset and the CIFAR-10 dataset, respectively. We observe that the learning performance of DAO-PPO is better than that of HE throughout the learning process but not as good as HO. This is because DAO-PPO does not completely eliminate local iteration disparity as HO does, leading to slightly inferior performance compared to HO. In the following experimental results on total latency and reward, we will demonstrate the advantages of using DAO-PPO.

\begin{figure}[htbp]
\centering
\vspace{-15pt}
\subfloat[Per-Round Total Latency with C-MAPSS]{\includegraphics[width=0.49\linewidth]{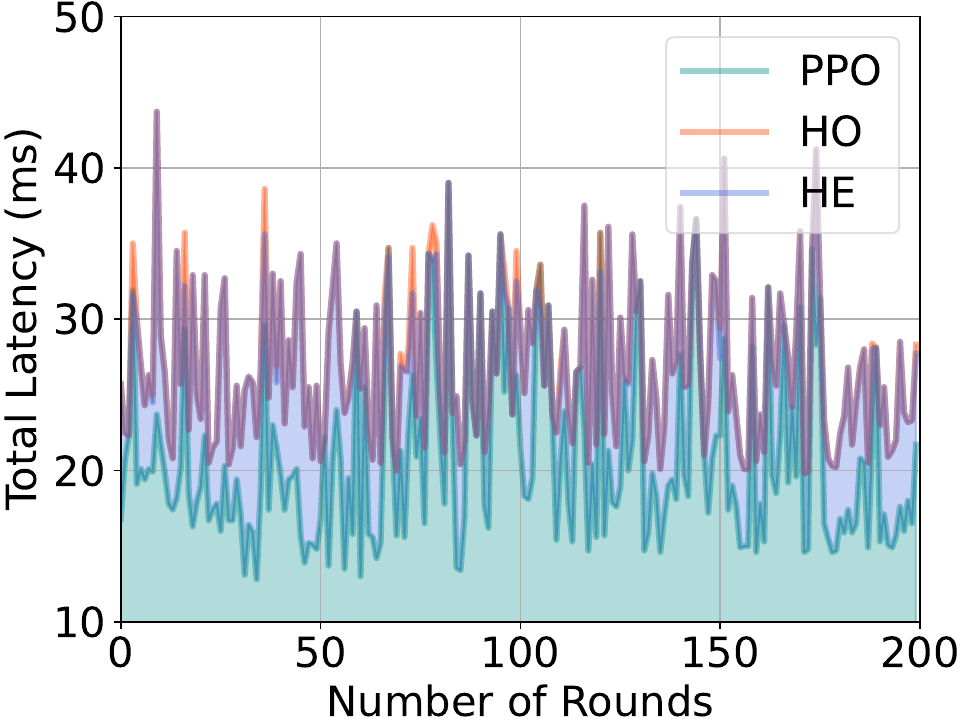}} 
\subfloat[Per-Round Total Latency with CIFAR-10]{\includegraphics[width=0.49\linewidth]{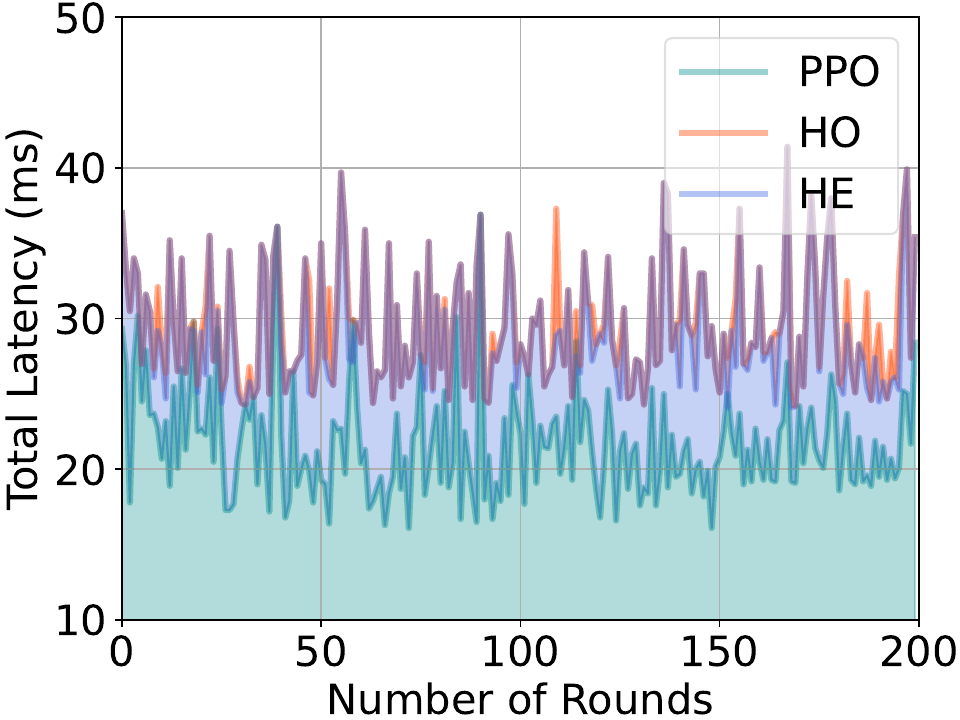}}\\
\subfloat[Average Total Latency with C-MAPSS]{\includegraphics[width=0.49\linewidth]{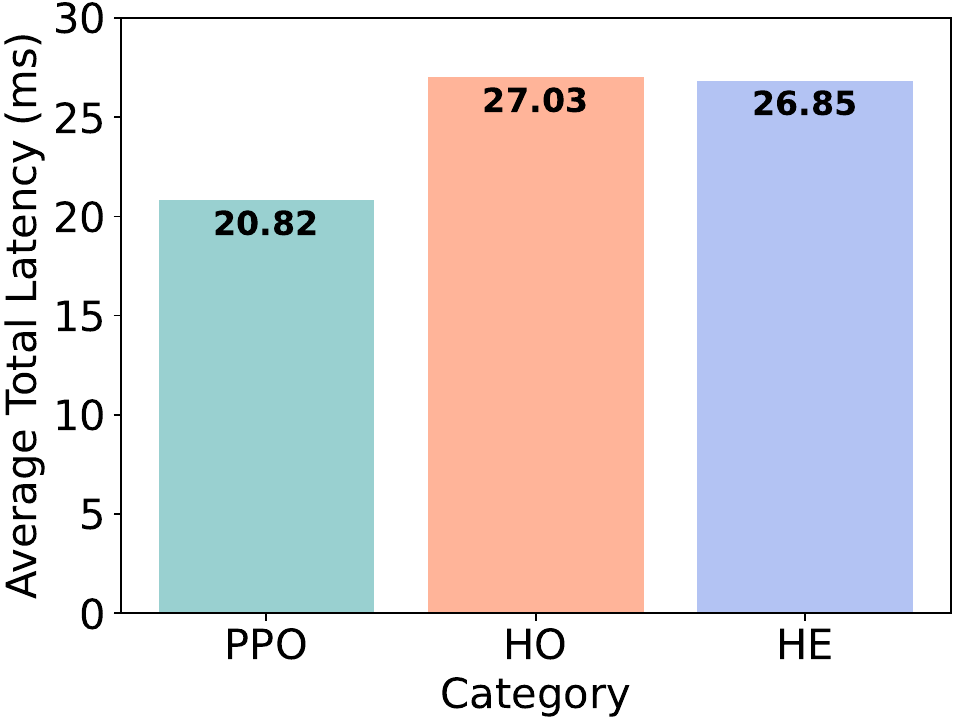}} 
\subfloat[Average Total Latency with CIFAR-10]{\includegraphics[width=0.49\linewidth]{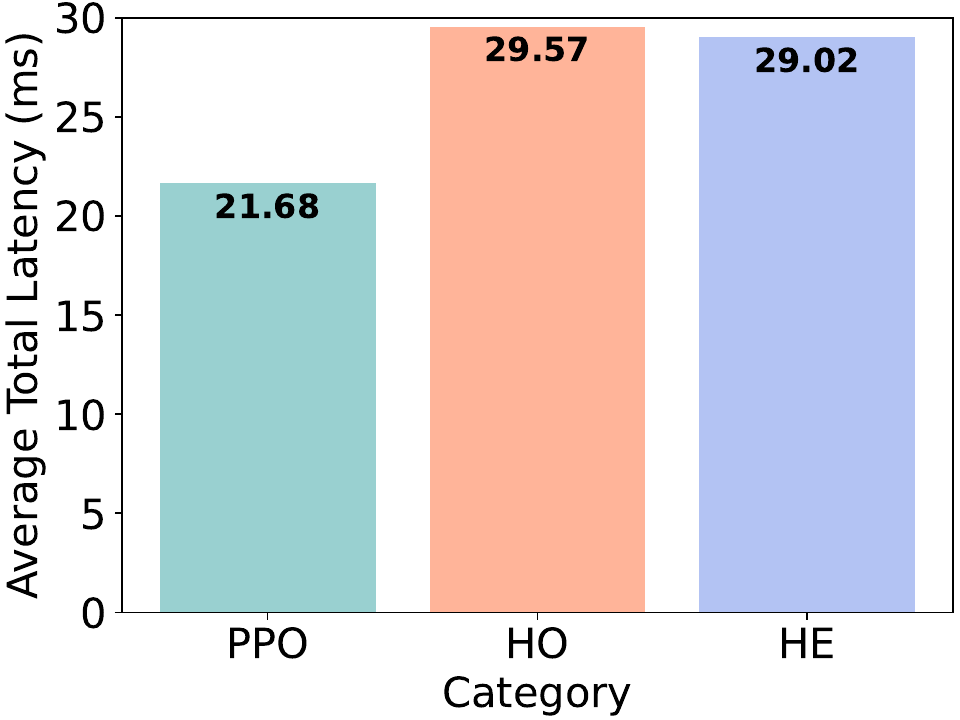}}
\caption{Comparison of Total Latency.}   \label{deonvfl-TL}
\vspace{-10pt}
\end{figure}

\textbf{Comparison of Total Latency}. In this section, we compare the total latency of DAO-PPO with benchmarks within a single run, focusing on two key metrics: per-round total latency and average total latency. Fig.\ref{deonvfl-TL}(a) and Fig.\ref{deonvfl-TL}(b) illustrate the per-round total latency for the C-MAPSS and CIFAR-10 datasets, respectively, while Fig.\ref{deonvfl-TL}(c) and Fig.\ref{deonvfl-TL}(d) present the average total latency for the C-MAPSS and CIFAR-10 dataset. The experimental results reveal two findings: first, DAO-PPO outperforms both HO and HE in terms of per-round and average total latency, demonstrating its effectiveness in reducing overall latency; second, HE slightly surpasses HO in per-round and average total latency, which can be attributed to occasional rounds where the sensor with the minimum local iteration decision avoids becoming a bottleneck, thereby reducing the overall latency.

\begin{figure}[htbp]
\centering
\vspace{-15pt}
\subfloat[Per-Round Reward with C-MAPSS]{\includegraphics[width=0.49\linewidth]{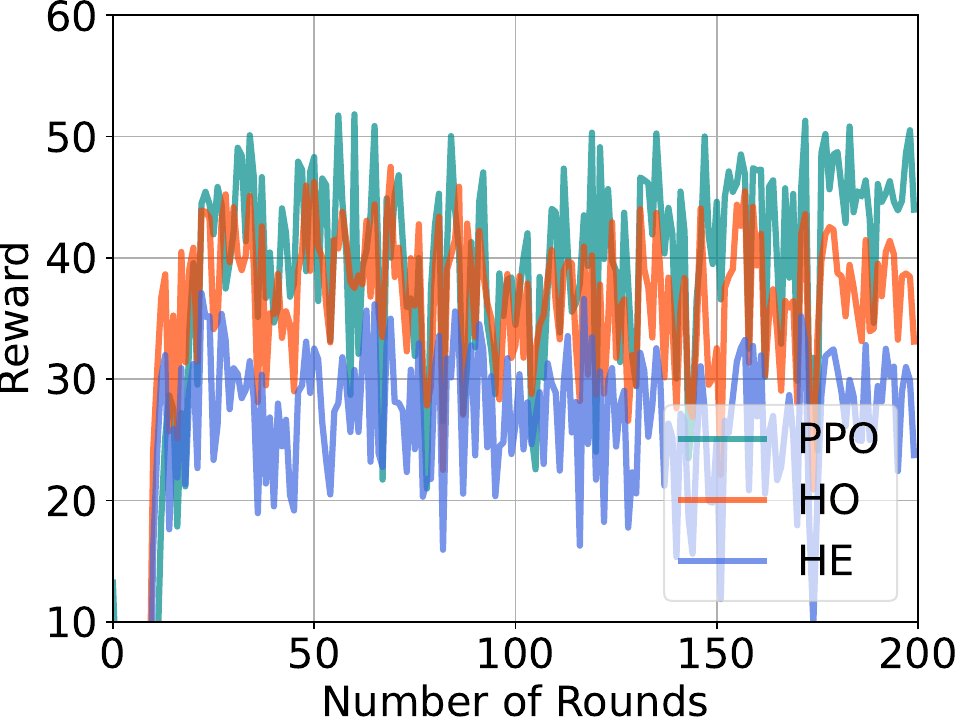}} 
\subfloat[Per-Round Reward with CIFAR-10]{\includegraphics[width=0.49\linewidth]{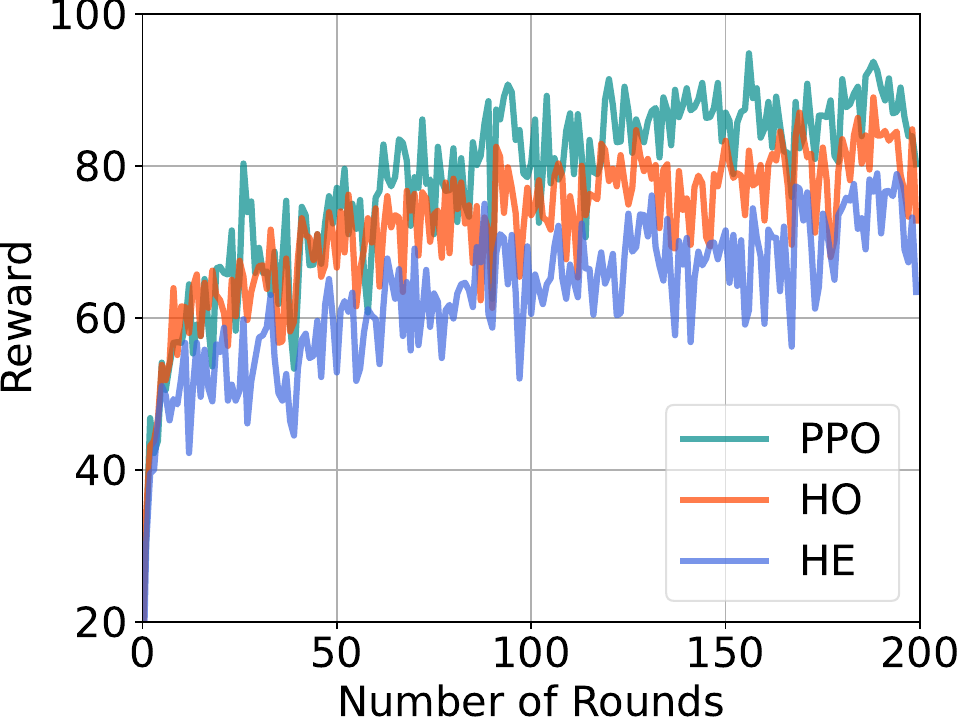}}\\
\subfloat[Average Reward with C-MAPSS]{\includegraphics[width=0.49\linewidth]{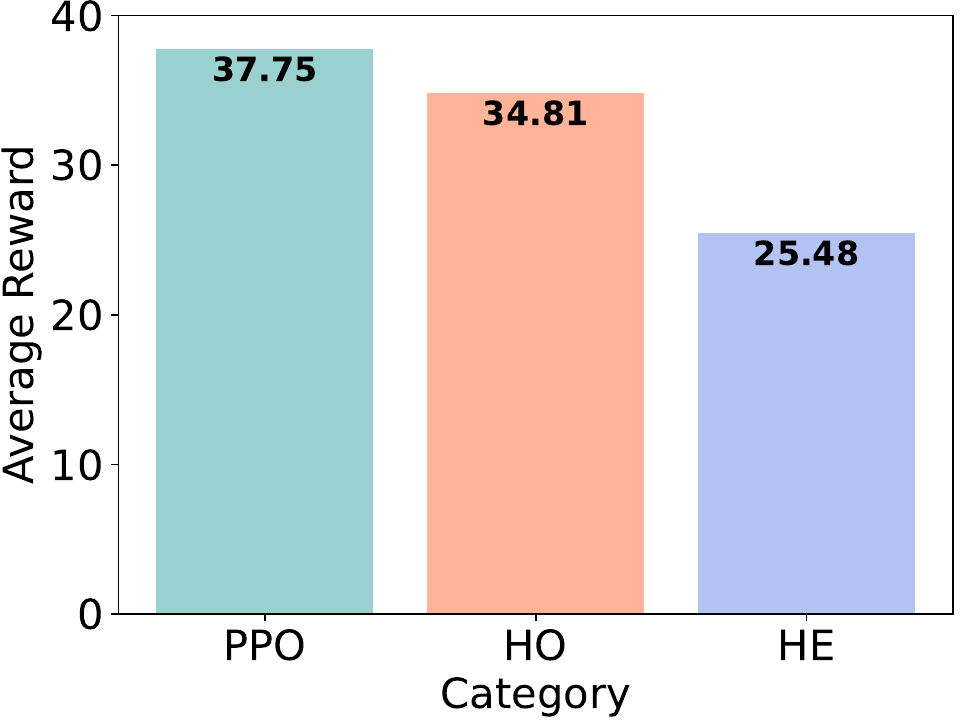}} 
\subfloat[Average Reward with CIFAR-10]{\includegraphics[width=0.49\linewidth]{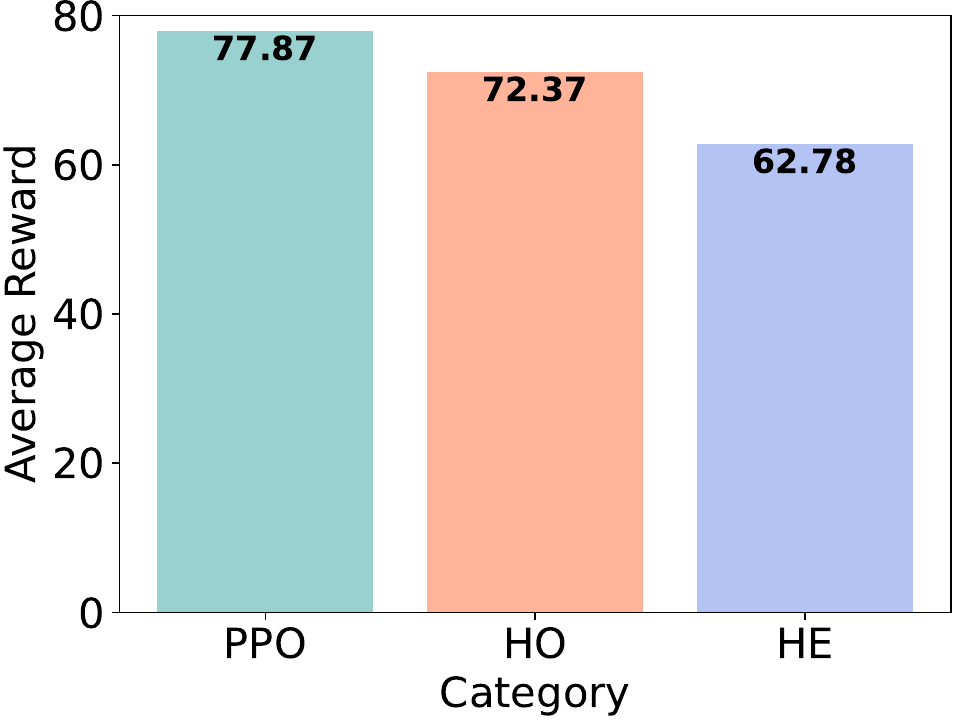}}
\caption{Comparison of Reward.}   \label{deonvfl-TR}
\vspace{-10pt}
\end{figure}

\textbf{Comparison of Reward}. In this section, we compare the reward of DAO-PPO with benchmarks within a single run. The evaluation focuses on two key metrics: per-round reward and average reward. Fig.~\ref{deonvfl-TR}(a) and Fig.~\ref{deonvfl-TR}(b) present the per-round reward for the C-MAPSS and CIFAR-10 dataset, while Fig.~\ref{deonvfl-TR}(c) and Fig.~\ref{deonvfl-TR}(d) show the average reward for the C-MAPSS and CIFAR-10 dataset. The results reveal two key findings: first, DAO-PPO achieves higher per-round and average rewards compared to HO and HE, as its learning objective is to maximize overall reward by effectively balancing learning performance, total latency, and local iteration disparity. Second, unlike the total latency results where HE and HO are relatively close, HO significantly outperforms HE in terms of reward. This is because HO is unaffected by local iteration disparity, whereas HE is impacted by it.

The experimental results above demonstrate the superiority of the proposed DAO-VFL algorithm over its corresponding benchmarks in both noise reduction and adaptive local iteration decision-making.

\section{Conclusion}
In this work, we proposed the DAO-VFL algorithm to address key challenges in real-world IIoT scenarios, specifically focusing on multi-sensor co-training framework for industrial assembly lines. In addition to the inherent challenge of integrating online learning with VFL, we tackled critical issues prevalent in industrial environments, such as communication noise and sensor heterogeneity. To mitigate the impact of communication noise, we employed a denoising autoencoder to reduce noise in feature embeddings during the training process. To address sensor heterogeneity, we leveraged deep reinforcement learning to determine local iteration decisions for each sensor as actions. This approach reduces total latency and local iteration disparity while maintaining robust learning performance. Our findings are supported by detailed theoretical analysis and extensive experimental validation. Future research will focus on the development of a testbed and optimizing performance in real-world industrial assembly line environments using actual industrial data, aiming to advance intelligent solutions for IIoT applications.

\bibliographystyle{IEEEtran}
\bibliography{bibligraphy}

\end{document}